\pdfoutput=1
\documentclass[runningheads]{llncs}

\usepackage{eccv}

\usepackage{eccvabbrv}
\usepackage{graphicx}
\usepackage{booktabs}
\usepackage[accsupp]{axessibility}
\usepackage{hyperref}
\usepackage{orcidlink}
\usepackage{overpic}
\usepackage{wrapfig}
\usepackage{tcolorbox}
\usepackage{multirow}
\usepackage{microtype}
\usepackage{booktabs}
\usepackage{tabularx}
\input{macros}
\begin{document}

\title{Reflecting Process Expertise in Procedural Material Generation}
\titlerunning{Reflecting Process Expertise in Procedural Material Generation}

\author{Kunal Gupta, Gaurav Joshi, Yen-Ru Chen, Seemandhar Jain, \\ Ishit Mehta and Manmohan Chandraker}
\authorrunning{K. Gupta et al.}

\institute{University of California San Diego, La Jolla, CA, USA}

\maketitle

\begin{figure*}[h]
\centering
\includegraphics[width=0.99\linewidth]{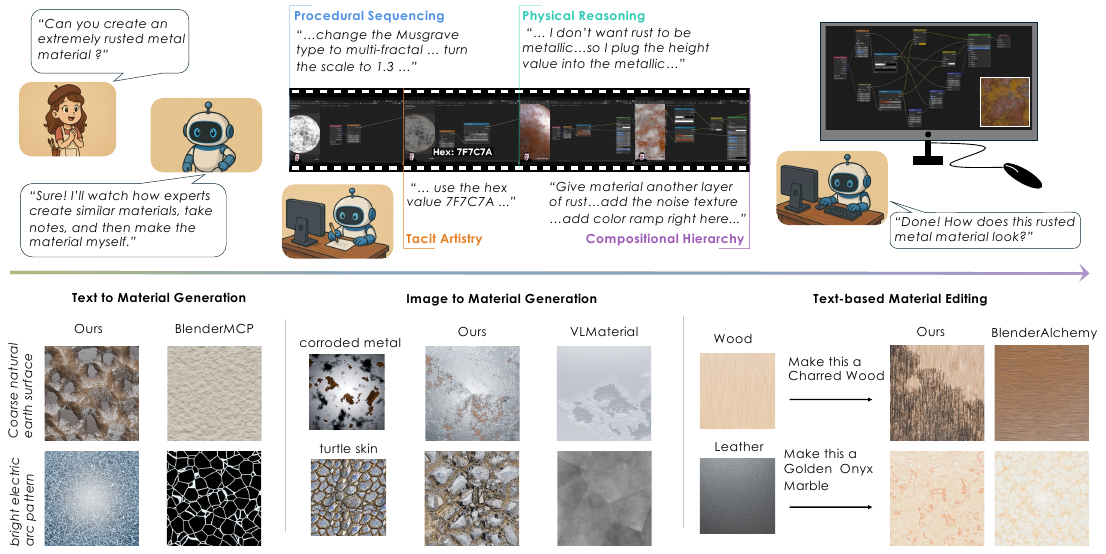}\\
\caption{\small \textbf{Retrieval-time process reasoning for procedural materials.}
(Top) Rather than synthesizing material graphs directly, our approach reasons over \emph{process traces} derived from expert demonstrations (e.g., tutorial videos), capturing procedural sequencing, physical intuition, and compositional design strategies used by artists. 
(Bottom) This process-centric reasoning enables diverse material creation tasks, including text-to-material generation (left), image-conditioned synthesis (center), and text-based editing (right). Across tasks, materials produced through process reasoning align more closely with user intent and exhibit stronger visual fidelity than prior graph-only methods such as BlenderMCP~\cite{blendermcp}, VLMaterial~\cite{livlmaterial}, and BlenderAlchemy~\cite{huang2024blenderalchemy}.}
\label{fig:teaser}
\vspace{-20pt}
\end{figure*}

\begin{abstract}
Procedural material creation underpins applications in digital content creation, visual effects, and 3D asset design. Achieving high-quality results requires more than reproducing node graphs—it demands understanding the process by which experts construct materials. We formulate procedural material generation as retrieval-time process reasoning over expert demonstrations, elevating process to a first-class representation beyond graph-only synthesis. Concretely, we represent expert workflows as \emph{process traces}: textual records of construction steps, parameters, and design intent. To instantiate this idea, we use a pretrained LLM-based \ps\ to synthesize a process trace aligned with a user’s intent and a pretrained LLM-based \com\ to ground the process trace into an executable Blender material graph. Because procedural expertise is most naturally conveyed through demonstrations, we leverage tutorial videos as a source of process knowledge and extract textual, LLM-compatible traces using automated video analysis tools. In an expert study with five Blender artists (avg. 7.5 years of experience), materials generated by reflecting expert demonstrations were found to produce workflows requiring fewer edits, and more closely match professional design strategies than methods operating solely on static artifacts. A user study with 150 participants further shows that our approach achieves superior generation and editing performance compared to prior procedural systems. All code, models, and data will be available at \url{https://materialapprentice.github.io}.
\vspace{-1em}
\end{abstract}

\section{Introduction}
\label{sec:introduction}

Procedural materials power the visual surfaces of films, games, product design,
and virtual worlds by encoding appearance as parameterized node graphs rather
than fixed bitmaps. Their flexibility makes them essential for scalable, editable,
and physically consistent asset creation. Yet crafting these materials remains
specialist work: experienced artists rely on visual intuition, procedural reasoning,
and domain-specific workflows. Both aspiring and established creators continually
refine their craft by observing how experts layer textures, tune parameters, and
reason about physical effects. Such demonstrations convey process knowledge that
extends far beyond the final graph. Figure~\ref{fig:teaser} illustrates our central
premise: material generation should reflect the process expertise artists use when
constructing procedural materials. Current methods, even when trained on large
procedural material datasets, operate in isolation at inference time: they generate
materials but do not incorporate such expertise into the generation.

\begin{figure*}[t]
    \centering
    \begin{overpic}[width=\textwidth]{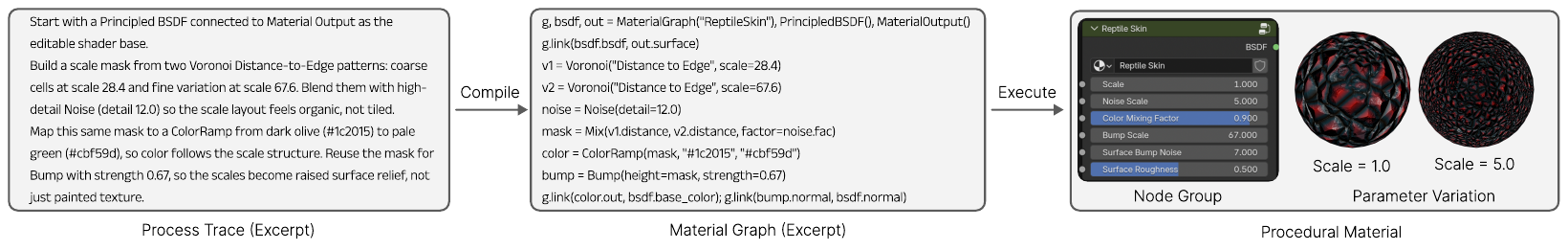}
    \end{overpic}
    \caption{\small \textbf{Three representations used in \ma.} A \emph{process trace} gives artist-like construction steps with concrete parameters and design intent. The trace is compiled into a \emph{material graph}, an executable shader structure. Executing the graph instantiates a \emph{procedural material} with editable controls and rendered appearances.}
    \label{fig:definition}
    \vspace{-2em}
\end{figure*}

If process knowledge is central to high-quality material design, how can it be
reflected automatically during generation? We address this by framing
\emph{procedural material generation as retrieval-time process reasoning}. We
introduce \ma, a multimodal framework that integrates expert-derived process
traces into the generative loop. Rather than operating solely in graph space, \ma\
retrieves relevant traces, synthesizes a target trace aligned with the user's intent,
and grounds it into an executable material graph. As illustrated in
Fig.~\ref{fig:definition}, \ma\ distinguishes three representations: a
\emph{process trace}, which records construction steps, parameters, and design
intent; a \emph{material graph}, the executable shader-node layout compiled from
the trace; and a \emph{procedural material}, the resulting editable shader asset.
Treating process traces as a first-class representation provides a foundation for
higher-quality, more editable material generation.

Expert procedural knowledge is typically acquired by observing skilled artists as they construct materials. In practice, this expertise is often externalized through tutorial videos, which provide a temporally ordered and multimodal record of the creation process: the evolving node graph, parameter adjustments, synchronized narration explaining intent, and visual focus on individual operations. These signals indicate not only the final artifact but the reasoning trajectory behind it.

In our setting, we approximate this observational learning by analyzing Blender tutorial videos and extracting structured, agent-friendly process traces using automated video analysis tools. Rather than forming a training dataset, these traces serve as demonstrations retrieved at inference time. In practice, only a small number of relevant exemplars is sufficient to guide synthesis, forming a lightweight knowledge base for retrieval-time reasoning.

Technically, we instantiate retrieval-time process reasoning through a retrieval-augmented large language model (LLM) framework. Given a user query, \ma\ retrieves relevant process traces and conditions a pretrained LLM-based \ps\ to synthesize a process trace aligned with the intended material. This trace is then grounded into an executable representation with a pretrained LLM-based \com\ (differing only in prompting), which translates it into a structured material graph implemented using Blender shader nodes. This design separates reasoning from execution while integrating process-level conditioning directly into the generative loop. Importantly, beyond one-shot synthesis, \ma\ can also \emph{demonstrate the process}, producing step-by-step constructions that mirror expert workflows and support structured editing. In contrast, prior systems such as VLMaterial~\cite{livlmaterial}, BlenderAlchemy~\cite{huang2024blenderalchemy}, and Blender-MCP~\cite{blendermcp} are limited to operation in graph space.


We evaluate \ma\ on a suite of Blender material creation tasks, including multimodal material generation and text-driven editing. Across quantitative metrics and perceptual studies, \ma\ consistently outperforms VLMaterial~\cite{livlmaterial}, BlenderAlchemy~\cite{huang2024blenderalchemy}, and Blender-MCP~\cite{blendermcp}. In addition, a study with experienced Blender artists shows that process traces generated by \ma\ more closely follow expert constructions and require fewer manual edits than alternatives.

In summary, we frame procedural material generation as retrieval-time process reasoning over expert demonstrations. Our contributions are:

\begin{itemize}
\item A formulation of procedural material generation as retrieval-time process reasoning, elevating process to a first-class representation beyond graphs.
\item A retrieval-augmented architecture with an explicit process-trace representation and compiler that separate reasoning from execution and ground generated traces into executable Blender shader graphs.
\item A curated corpus of Blender tutorial videos and video analysis tools for extracting agent-friendly process traces that enable inference-time conditioning.
\item Comprehensive evaluation, including with Blender artists, showing process-based synthesis better aligns with expert workflows than graph-based systems.
\end{itemize}

\section{Related Work}
\label{sec:related}


\paragraph{Generative Materials}
Recent work directly synthesizes material appearances. MaterialGAN~\cite{guo2020materialgan}
extends StyleGAN2~\cite{karras2020stylegan2} to generate SVBRDF textures, but
operates in an opaque latent space. Diffusion-based methods improve visual quality:
ControlMat~\cite{vecchio2024controlmat} uses ControlNet-guided diffusion for
SVBRDF generation with limited structural control, DreamMat~\cite{zhang2024dreammat}
maps text prompts to point-wise BRDF parameters in a hash grid, and
MaterialPicker~\cite{ma2025materialpicker} fine-tunes a video generator for
text- and image-conditioned material textures. Huang et al.~\cite{huang2025material}
generate textures directly in UV space, improving surface consistency. However,
these methods produce non-procedural, sample-based or latent representations and
do not expose editable procedural graphs.

\vspace{-0.2cm}
\paragraph{Generative Procedural Materials} Closest to our work are methods that synthesize procedural material graphs. Hu et al.~\cite{hu2023generating} generate node-based materials conditioned on
text prompts, example images, or partial node graphs, using node graphs as
defined in Adobe Substance 3D~\cite{adobe2023} and the Matformer
representation~\cite{guerrero2022}. Li et al.~\cite{li2024procedural} build on
these ideas, using synthetic data to train a transformer model for material
graph generation and applying RL post-training on both real and synthetic data
to improve realism and stability. While these methods produce procedural
representations that are more editable and interpretable than purely
texture-based outputs, they rely on large-scale training and custom models.

\vspace{-0.2cm}
\paragraph{LLMs for Program Synthesis} Large language models (LLMs) have demonstrated strong capabilities in program
synthesis~\cite{chen2021evaluating}, motivating their use as engines for
3D content creation. Early systems directly generate Python code for Blender,
such as 3DGPT~\cite{sun20253d}. However,
operating in a general-purpose, low-level language makes it difficult to
maintain structure, guarantee validity, and scale to complex scenes.
To address these limitations, several works introduce domain-specific languages
(DSLs) tailored for 3D generation. Holodeck~\cite{yang2024holodeck} and
L3GO~\cite{yamada2025l3go} use LLMs to author scenes in compact DSLs,
simplifying high-level planning and constraint reasoning. While DSLs can make
LLM outputs more structured, they typically lack large-scale training corpus, which can lead to unexpected behavior~\cite{barbero2024evaluation}.

\vspace{-0.2cm}
\paragraph{LLMs as Planners} Another line of work treats LLMs as planners that orchestrate external tools
for 3D creation. LayoutGPT~\cite{feng2024layoutgpt} uses LLMs to generate layouts, while SceneCraft~\cite{hu2024scenecraft}
and LayoutVLM~\cite{sun2025layoutvlm} adopt an analysis-by-synthesis loop in
which a vision-language critic evaluates rendered scenes and guides refinement.
When combined with tool-use frameworks like ReAct~\cite{yao2022react}, LLM
agents can scale to interactive, photorealistic environments~\cite{yao2025cast}.

\vspace{-0.2cm}
\paragraph{VLMs for Materials} LLMs and vision-language models have also been applied directly to material
synthesis. VLMaterial~\cite{livlmaterial} fine-tunes a VLM to generate
procedural materials represented as node graphs and
BlenderAlchemy~\cite{huang2024blenderalchemy} iteratively samples and refines
node graphs using VLM-based evaluation of rendered outputs, demonstrating that large models can reason about procedural material programs and
visual feedback. In contrast, our method is training-free, leverages
process-trace reasoning, and uses in-context learning from expert material
generation workflows to achieve high-quality procedural material synthesis and
editing.
\vspace{-2em}

\section{Method}
\label{sec:method}
\begin{figure*}[t]
    \centering
    \begin{overpic}[width=\textwidth]{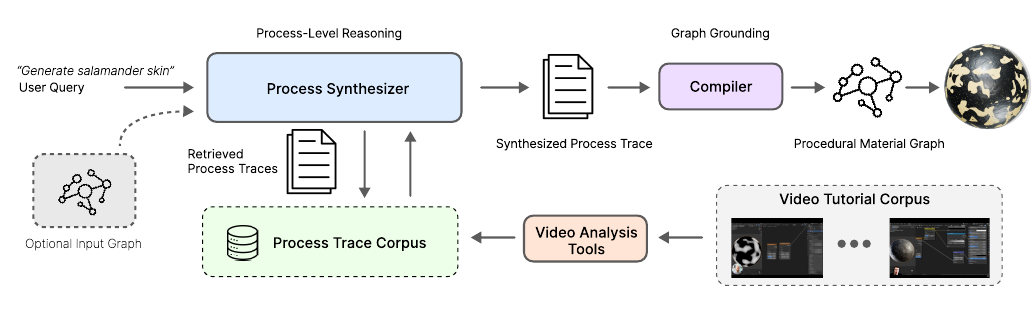}
    \end{overpic}
    \vspace{-0.7cm}
    \caption{\small \textbf{Overview:}
Given a user query and optionally an input graph, the \ps\ retrieves relevant
process traces from a corpus of expert traces and synthesizes a target process
trace aligned with the query. These traces are extracted automatically from
tutorial videos using video analysis tools that convert demonstrations into
textual records of construction steps, parameters, and design intent. The \com\
then grounds the synthesized trace into an executable Blender procedural
material graph.}
    \vspace{-0.3cm}
    \label{fig:pipeline}
    \vspace{-0.3cm}
\end{figure*}

\vspace{-0.7em}
Procedural material creation is not merely graph assembly, but the execution of a
structured reasoning trajectory. Skilled artists iteratively refine structure, adjust
parameters, and articulate intent; the final node graph is only the endpoint of this
process. We therefore distinguish the three representations used by \ma: a
\emph{process trace}, which records construction steps, parameters, and design
intent; a \emph{material graph}, the executable shader-node layout compiled from
the trace; and a \emph{procedural material}, the resulting editable shader asset.
While prior systems operate directly in graph space, we perform generation in
process space.

\begin{figure*}[t]
    \centering
    \begin{overpic}[width=\textwidth]{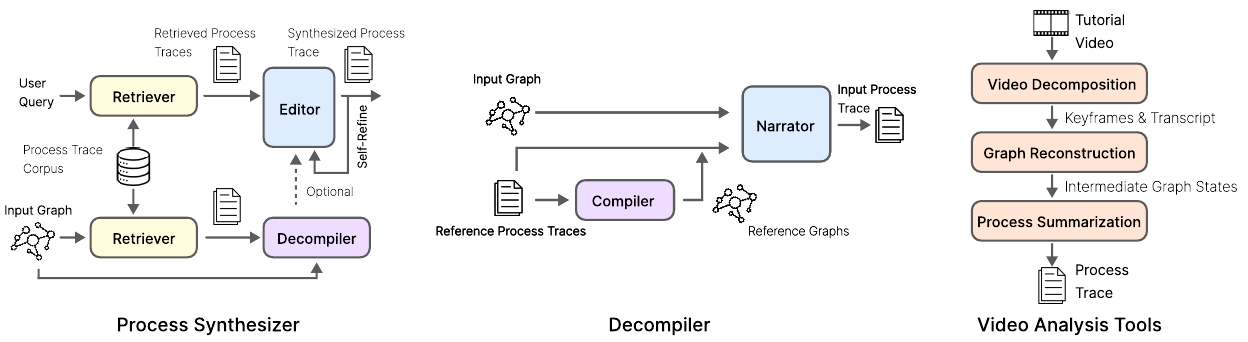}
    \end{overpic}
    \vspace{-0.4cm}
    \caption{\small \textbf{Components of \ma. }
(Left) \ps: Given a user query, the retriever selects relevant process
traces, which condition the \editor\ to synthesize a target trace. For editing,
an input graph is first converted into a process trace by the \dcom.
(Middle) \dcom: An input material graph is converted into a process
trace using exemplar graph--trace pairs; the \narrator\ matches graph
substructures to similar patterns in the exemplars and transfers their associated
construction steps.
(Right) Video analysis tools: Tutorials are decomposed into keyframes
and transcripts, intermediate graph states are reconstructed, and multimodal
signals are summarized into process traces for the \ps.
    }
    \label{fig:components}
    \vspace{-0.3cm}
\end{figure*}

To operationalize this formulation, \ma\ instantiates retrieval-time process reasoning. As summarized in Figure~\ref{fig:pipeline}, given a user query, it retrieves relevant process traces and synthesizes a process trace via \ps\ aligned with the intended material. This trace is then grounded into an executable Blender shader graph through the \com, explicitly separating reasoning from execution. Process traces are derived from expert video demonstrations using video analysis tools which convert raw tutorials into structured, agent-friendly textual representations. These traces form the knowledge base for retrieval-time reasoning. Figure~\ref{fig:components} details the \ps, \dcom, and video analysis tools, which are described in the following subsections.
\vspace{-1em}

\subsection{Process Synthesizer}
\label{sec:ps}
The \ps\ is the core reasoning module of \ma, synthesizing process traces aligned
with user intent. Given a query $q$, a \retriever\ selects relevant reference traces $P$ from the demonstration corpus $C_P$. Conditioned on $q$ and $P$, the \editor\ synthesizes a target trace $p$ describing how to construct the desired
material.

For editing, \ps\ additionally accepts an input graph $g_i$. In this setting, the \retriever\ selects process traces relevant to $g_i$, and a \dcom\ module reconstructs a corresponding process trace $p_i$. This inferred process trace, together with the user’s edit instruction, conditions the \editor\ to synthesize an updated process trace consistent with the requested modification.
\vspace{-0.7em}
\paragraph{Retriever.}
The \retriever\ selects relevant demonstrations for both generation and editing. Rather than ranking examples by surface appearance alone, it uses the pretrained LLM to compare how materials are constructed. Given material names and descriptions in $C_P$, it predicts which expert demonstrations best match the user's intent, desired material properties, and construction strategy. The top-$K$ corresponding process traces are selected to condition synthesis.
\vspace{-0.7em}

\begin{figure*}[t]
    \centering
    \begin{overpic}[width=\textwidth]{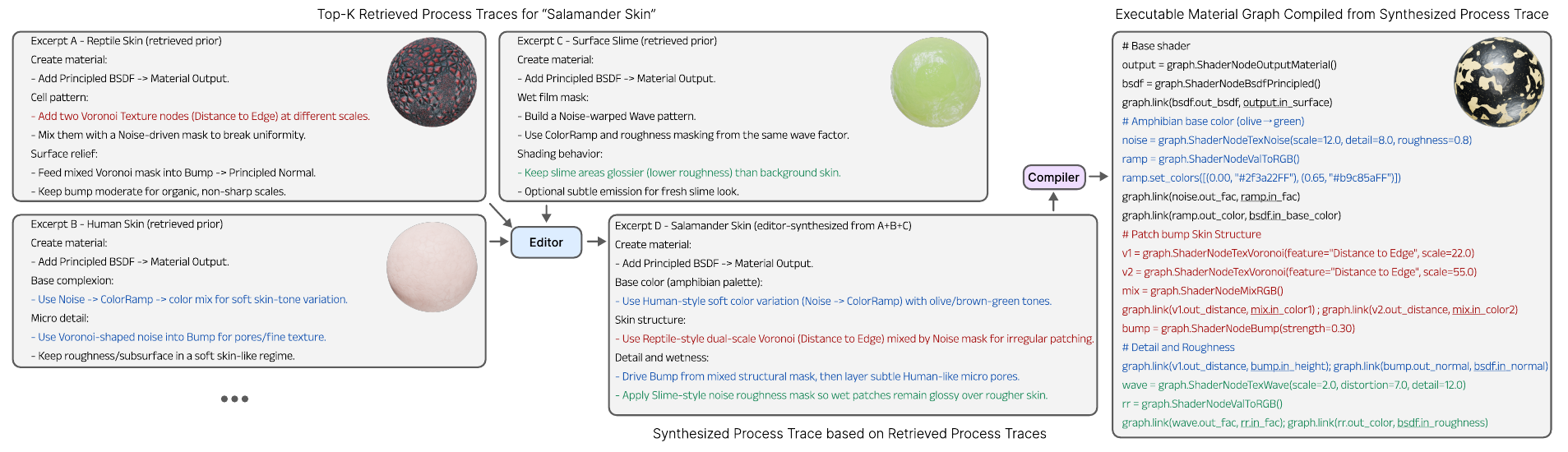}
    \end{overpic}
    \caption{\small \textbf{Process-trace synthesis by reflecting on retrieved demonstrations.}
(Left) Top-$K$ retrieved process traces for the query ``salamander skin''.
Although visually distinct, these demonstrations contain useful construction
patterns---scale structure from reptile skin, fine bump detail from human skin,
and smooth organic noise from surface slime.
(Center) The \editor\ uses these retrieved traces to synthesize a target
process trace that combines their construction strategies for the query.
(Right) The \com\ translates the synthesized trace into an executable
material graph program, which produces the final material.
    }
    \vspace{-0.3cm}
    \label{fig:process_trace_construction}
    \vspace{-0.3cm}
\end{figure*}

\paragraph{Editor.}
The \editor\ is a pretrained LLM prompted to perform in-context synthesis in the
\emph{process} domain. Conditioned on $P$ and $q$, it generates a target process
trace $p$ specifying the construction sequence, parameter choices, and compositional
structure required to realize the requested material. As illustrated in
Fig.~\ref{fig:process_trace_construction}, the \editor\ combines construction patterns from
retrieved traces into a target trace before compilation. Here, lowercase $p$ denotes
a single trace, while $P$ refers to sets of retrieved process traces from $C_P$.
To improve reliability, the \editor\ employs a lightweight self-refinement step,
critiquing and revising its draft before finalizing $p$, enhancing consistency and
adherence to the query. Because the \editor\ operates purely in process space,
editing requires representing existing graphs in the same modality, motivating the
\dcom\ module.
\vspace{-0.7em}
\paragraph{DeCompiler.}
The \dcom\ converts an input material graph $g_i$ into a process trace, enabling
editing within a unified representation. Given $g_i$, it reconstructs a plausible
trace $p_i$ using exemplar graph--trace pairs from the reference corpus. Each
reference trace is first compiled into graph form via the \com, yielding exemplar
pairs $E = \{(g_j, p_j)\}_{j=1}^k$. The exemplars are selected to match the input
graph in construction strategy rather than surface appearance. A specialized LLM
module, \narrator, then reconstructs
$p_i = \narrator(g_i; E)$
by matching substructures of $g_i$ to similar graph patterns in the exemplars and
transferring their associated construction steps. The resulting $p_i$ captures the
operations, parameters, and intent implied by the input graph.
\vspace{-1em}

\subsection{Grounding Process Traces in Material Graphs}
\label{sec:com}

The \com\ translates textual process traces into executable Blender shader graphs. It is implemented as a pretrained LLM prompted to generate Blender Python graph-construction code conditioned on a trace $p$. A lightweight self-refinement stage enforces consistency between node topology, parameter definitions, and layer ordering. Given $p$, the compiler produces an executable graph $g$. The \com\ generates multiple candidate programs $N$ per synthesized process trace and selects the first syntactically valid and executable instance. The compiler targets Blender versions 3.0–4.5, with custom node groups for compatibility across API changes.
\vspace{-1em}

\subsection{Video Tutorials for Extracting Process Traces}
\label{sec:tutorial_corpus}

To support retrieval-time process reasoning, we construct a corpus of expert Blender material tutorials. The corpus consists of 158 publicly available YouTube tutorials from established creators, selected for clear stepwise instruction, synchronized narration, and focused visual presentation (e.g., zoomed-in node editing, explicit parameter explanation, and structured workflow progression). It spans 29 material categories (e.g., metals, fabrics, organic surfaces), with an average of 5.5 tutorials for each. Tutorials average 12.6 minutes in duration, with materials containing on average 14.3 shader nodes, providing rich signals about compositional structure, parameter tuning, and procedural reasoning.

In addition, we create 10 original tutorials for which we hold full rights and will release publicly to ensure long-term reproducibility. Publicly sourced tutorials are used exclusively as a retrieval-time knowledge source; no model training is performed on raw video content. We release YouTube IDs and category annotations to enable reconstruction of the corpus while preserving attribution to original creators. Detailed statistics are provided in the supplementary.
\vspace{-1em}
\subsection{Process Extraction via Video Analysis Tools}
\label{sec:video_tools}

To derive process traces from expert demonstrations, \ma\ uses automatic video
analysis tools to convert tutorials into structured text that LLM agents can
directly retrieve and reason over. This component is primarily an engineering step rather than a core research contribution: the goal is simply to automatically extract procedural signals from demonstrations so that they can be used for retrieval-time reasoning. In practice, Blender tutorials exhibit a simple box-and-wire interface, allowing standard detection and segmentation models to reliably recover graph structure. Given a tutorial video $v$, the pipeline extracts keyframes, reconstructs intermediate graph states, and aligns visual edits with narrated explanations.

\vspace{-0.7em}
\paragraph{Video Decomposition.}
Videos are first decomposed into frames and audio narration. Informative keyframes are selected to capture meaningful transitions in the material construction process while removing redundant views. The audio stream is transcribed to preserve the artist’s verbal explanations and design rationale.

\vspace{-0.7em}
\paragraph{Graph Reconstruction.}
The visual stream is analyzed to recover the evolving node graph. From selected keyframes, the system identifies node layouts, connectivity patterns, and parameter states using standard detection and segmentation models fine-tuned for the Blender shader editor. This produces a sequence of intermediate graph representations reflecting the progressive assembly of the material.

\vspace{-0.7em}
\paragraph{Process Summarization.}
A \summarizer\ module integrates the reconstructed graph states with the transcribed narration to produce a compact textual process trace $p_v$, capturing both the sequence of construction steps and the accompanying procedural reasoning expressed by the artist. These traces populate the retrieval corpus used by the \ps\ at inference time.

\vspace{-0.7em}
\paragraph{Robustness of Process Extraction.}
Three design choices make extraction reliable. (1) \emph{Grounding}: frame understanding relies on explicit vision tasks -- node, socket, and wire detection -- rather than direct LLM inference. Given the structured box-and-wire layout of the Blender shader editor, these detectors achieve mAP50 = 0.98 and mIoU = 0.87. (2) \emph{Selection}: a fine-tuned frame-selection model retains only visually parseable frames robust to zoom, pan, and occlusion. (3) \emph{Redundancy}: frames are parsed at 0.2 Hz and combined with narration through an LLM summarizer, mitigating localized perception errors and producing stable process traces. Further, while the pipeline contains multiple stages, errors in early stages are mitigated by the process abstraction. The \ps\ reasons over construction patterns rather than exact node configurations, allowing it to tolerate minor inaccuracies in reconstructed graph states or narration alignment. In practice, we observe that small parsing errors rarely propagate to catastrophic failures in the final material graph. Further details are provided in the supplementary.


\vspace{-1em}

\subsection{Implementation Details}
\label{sec:implementation}

All LLM-based components—\retriever, \editor, \dcom, \com, and \summarizer—use the same pretrained GPT-5 model, differing only in prompt structure and contextual conditioning. The \retriever\ selects the top $K=3$ process traces per query, balancing contextual coverage and computational efficiency. The compiler generates $N=10$ programs. While processes synthesized by the \ps\ are typically coherent and faithful to the query, final material quality may be further refined through established procedural optimization techniques~\cite{huang2024blenderalchemy}. A single process trace is $\approx 8k$ tokens, a single run costs $\approx 2$USD and runs for $30$ mins.
\vspace{-1em}
\section{Experiments}
\label{sec:experiments}
Our experiments probe the implications of viewing procedural material generation as \emph{retrieval-time process reasoning}. 
Section~\ref{sec:process_reasoning} evaluates the key design choices implied by this formulation—reasoning in process space rather than graph space and grounding execution through the \com. 
Section~\ref{sec:baseline_comp} then situates \ma\ within existing procedural material systems through comparisons on established generation and editing benchmarks. 
Finally, Section~\ref{sec:ablations} presents ablations analyzing generalization, retrieval behavior, the impact of video analysis tools, and the reproducibility of our framework through experiments on in-house tutorial videos.

\subsection{Reasoning in the Process Domain}
\label{sec:process_reasoning}

\begin{table*}[t!]
\centering
\small
\setlength{\tabcolsep}{3.5pt}
\renewcommand{\arraystretch}{1.08}
\resizebox{\textwidth}{!}{%
\begin{tabular}{@{}lcc@{\hspace{0.8em}}lc@{\hspace{0.8em}}lc@{}}
\toprule
\multicolumn{3}{c}{\textbf{Edit Effort}} &
\multicolumn{2}{c}{\textbf{Expert Preference}} &
\multicolumn{2}{c}{\textbf{Editing Sentiment}} \\
\cmidrule(lr){1-3}
\cmidrule(lr){4-5}
\cmidrule(l){6-7}
\textbf{Metric} & \textbf{Proc.} & \textbf{Graph} &
\textbf{Criterion} & \textbf{Pref.} &
\textbf{Signal} & \textbf{$\Delta$(P-G)} \\
\midrule
Node edits $\downarrow$    & \textbf{2.42}  & 3.42
& Impl. clarity $\uparrow$  & \textbf{100\%}
& Confusion $\downarrow$    & \textbf{-0.15} \\

Conn. edits $\downarrow$   & \textbf{5.42}  & 7.42
& Proc. strategy $\uparrow$ & \textbf{83\%}
& Hesitation $\downarrow$   & \textbf{-0.10} \\

Param. tweaks $\downarrow$ & 5.75           & \textbf{4.83}
& Outcome quality $\uparrow$ & \textbf{92\%}
& Satisfaction $\uparrow$   & \textbf{+0.40} \\

Struct. edits $\downarrow$ & \textbf{7.84}  & 10.84
& Param. control $\uparrow$ & \textbf{83\%}
& Completion $\uparrow$     & \textbf{+0.25} \\

Total edits $\downarrow$   & \textbf{13.58} & 15.67
& Pedagogy $\uparrow$       & \textbf{84\%}
&                           & \\

                            &                &
& Final render $\uparrow$   & \textbf{83\%}
&                           & \\
\bottomrule
\end{tabular}%
}
\caption{\small
\textbf{Expert Study.}
(Left) Average refinement edits.
(Middle) LLM-assigned preferences from artists' think-aloud commentary; final render preference is provided by artists after refinement.
(Right) Process$-$Graph sentiment deltas after mapping low/medium/high to 1/2/3.
Process-derived graphs require fewer structural/total edits, are preferred on key criteria, and reduce editing friction. 
}
\label{tab:expert_workflow_analysis}
\vspace{-2em}
\end{table*}
\begin{figure}
    \centering
    \includegraphics[width=\linewidth]{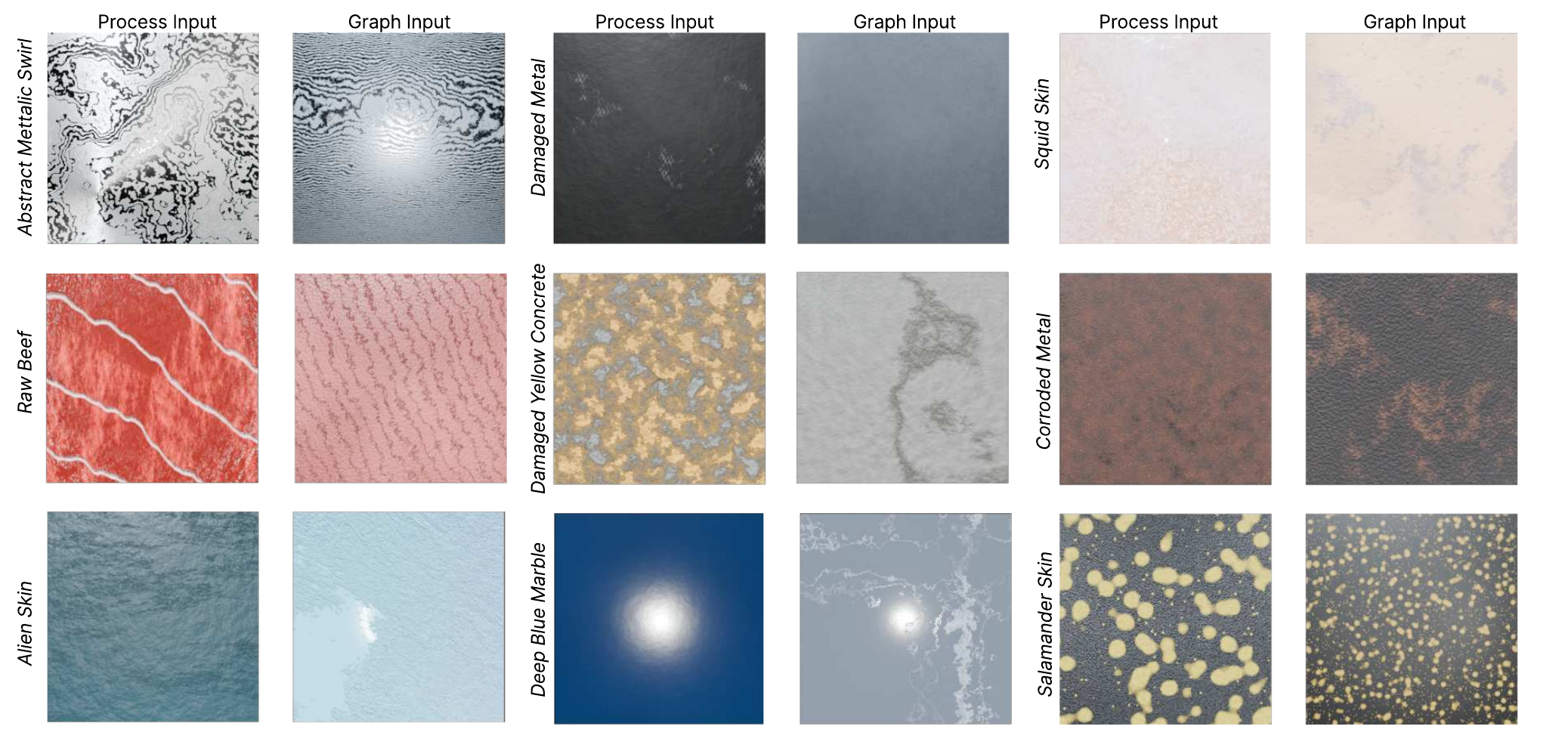}
    \caption{\small \textbf{Qualitative comparison of fully automatic generation using process-conditioned and graph-conditioned retrieval.} Each pair shows outputs from the complete pipeline without manual editing. Process-conditioned generation better captures the intended procedural structure and visual characteristics, while graph-conditioned retrieval often produces oversimplified or structurally inconsistent patterns.}
    \label{fig:auto_generation_pref}
    \vspace{-2em}
\end{figure}
To evaluate process-space reasoning, we isolate the representation used for retrieval while keeping the output modality fixed. Our method retrieves \emph{process traces}, while the baseline retrieves \emph{material graphs}; both variants synthesize process traces that artists follow to construct and refine material graphs. This isolates the effect of structured procedural knowledge relative to raw graph structure.

\vspace{-0.7em}
\paragraph{Evaluation setup.}
We recruit five Blender artists (avg. 7.5 years of procedural-material experience) and evaluate 60 synthesized process traces from 30 prompts under two retrieval conditions. For each prompt, artists see anonymized, randomly ordered process traces from the two variants. The study has two stages: artists first construct a material graph by following each trace, then refine the graph for up to ten minutes to better match the input prompt. We record atomic graph edits---node edits, connection edits, and parameter refinements---as a direct measure of corrective editing effort.

Artists also follow a think-aloud protocol\cite{ericsson1980verbal} during construction and refinement, verbalizing their understanding of each trace, the rationale behind their edits, and their assessment of the resulting graph. We ask them to comment on implementation clarity, procedural strategy, expected outcome quality, parameter control, and pedagogical value. An LLM then analyzes the recorded narrations with a fixed rubric to assign structured pairwise preferences and editing-friction labels. Final render preference is collected separately after refinement. Table~\ref{tab:expert_workflow_analysis} summarizes the study along three axes: edit effort, expert preference, and think-aloud sentiment. We provide the GPT-5 analysis prompts in the supplement.

\vspace{-0.7em}
\paragraph{Editing effort.}
The left block of Table~\ref{tab:expert_workflow_analysis} reports edits required to refine the generated graphs. Node and connection edits modify graph structure and therefore reflect changes to the underlying procedural strategy, while parameter refinements tune existing controls such as sliders and color ramps. Process-conditioned traces require fewer structural edits than graph-conditioned traces (7.84 vs.~10.84) and fewer total edits overall (13.58 vs.~15.67), indicating that they produce graphs closer to the intended procedural structure. They require slightly more parameter tweaks (5.75 vs.~4.83), which we interpret as a favorable shift: artists spend less effort repairing structure and more effort tuning exposed controls.

\vspace{-0.7em}
\paragraph{Expert preference.}
The middle block reports LLM-assigned preferences from artists' think-aloud commentary; final render preference is provided directly by artists after refinement. Process-conditioned workflows are favored across all criteria, especially implementation clarity (100\%) and outcome quality (92\%), with strong preferences for procedural strategy (83\%), parameter control (83\%), pedagogical value (84\%), and final render quality (83\%). This indicates that process traces provide clearer construction guidance beyond final appearance.

\vspace{-0.7em}
\paragraph{Editing sentiment.}
The right block analyzes editing friction from think-aloud narrations. An LLM assigns low/medium/high labels to four signals: \emph{confusion} and \emph{hesitation}, which indicate difficulty understanding the graph or choosing the next edit, and \emph{satisfaction} and \emph{completion}, which indicate successful refinement and confidence in the result. After mapping labels to 1/2/3, process-conditioned workflows reduce confusion ($-0.15$) and hesitation ($-0.10$), while increasing satisfaction ($+0.40$) and completion ($+0.25$). This suggests that process-derived graphs are easier for artists to understand, modify, and finalize.

\vspace{-1em}
\paragraph{Fully automatic generation.}
Figure~\ref{fig:auto_generation_pref} evaluates the complete pipeline without manual editing. For 50 diverse material prompts, we compare paired outputs from process-conditioned and graph-conditioned retrieval. Three experts judge each pair against the input prompt, with majority vote determining preference. Process-conditioned outputs are preferred in 72\% of comparisons, showing that process retrieval improves not only artist-facing editability but also automatic prompt alignment. Qualitatively, process-conditioned retrieval better captures hierarchical procedural structure, including directional fibers, layered corrosion, and structured surface noise, whereas graph-conditioned retrieval often yields simpler or structurally incorrect patterns.

\vspace{-1em}
\paragraph{Need for Compiler.}
We evaluate a variant where the \editor\ directly generates material graphs instead of processes. We measure reliability by \emph{execution rate}: the fraction of prompts for which at least one candidate program runs successfully in Blender. Across 100 prompts, direct graph generation fails in 30\% of cases, whereas our full process-to-compiler pipeline achieves a 100\% execution rate.

\vspace{-1em}

\subsection{Comparison with Prior Procedural Material Systems}

\begin{table}[t]
\centering
\small
\renewcommand{\arraystretch}{1.08}
\setlength{\tabcolsep}{3pt}
\resizebox{\textwidth}{!}{%
\begin{tabular}{llccccc@{\hskip 6pt\vrule\hskip 6pt}llccc}
\toprule
\multicolumn{7}{c@{\hskip 6pt\vrule\hskip 6pt}}{\textbf{Image-to-Material Generation}} &
\multicolumn{4}{c}{\textbf{Text-based Material Editing}} \\
\midrule
\textbf{Dataset} & \textbf{Method} &
\textbf{CLIP $\downarrow$} &
\textbf{Style Loss $\downarrow$} &
\textbf{SWD $\downarrow$} &
\textbf{GPT-5 Pref. $\uparrow$} &
\textbf{User Pref. (\%) $\uparrow$} &
\textbf{Dataset} & \textbf{Method} &
\textbf{CLIP $\downarrow$} &
\textbf{GPT-5 Pref. $\uparrow$} &
\textbf{User Pref. (\%) $\uparrow$} \\
\midrule
\multirow{4}{*}{\textbf{BlenderKit}}
& VLMaterial         & \textbf{0.856} & \textbf{0.040} & \textbf{2.44}  & 0.660 & 80  &
\multirow{4}{*}{\textbf{Coarse}}
& BlenderAlchemy      & 0.244          & 0.611 & 85  \\
& BlenderMCP         & 0.781          & 0.061          & 4.428          & 0.835 & 86  &
& BlenderMCP          & \textbf{0.237} & 0.611 & 80  \\
& Nearest Neighbour  & 0.791          & 0.059          & 4.326          & 0.874 & 84  &
& Nearest Neighbour   & 0.238          & 0.722 & 100 \\
& Ours               & 0.852          & 0.043          & 2.567          & --    & --  &
& Ours                & 0.260          & --    & --  \\
\midrule
\multirow{4}{*}{\textbf{In the Wild}}
& VLMaterial         & 0.764          & 0.066          & 8.973          & 0.771 & 89  &
\multirow{4}{*}{\textbf{Fine-grained}}
& BlenderAlchemy      & 0.240          & 0.643 & 81  \\
& BlenderMCP         & 0.737          & 0.080          & 8.227          & 0.800 & 85  &
& BlenderMCP          & \textbf{0.215} & 0.929 & 100 \\
& Nearest Neighbour  & 0.748          & 0.071          & 7.373          & 0.714 & 87  &
& Nearest Neighbour   & 0.228          & 0.786 & 100 \\
& Ours               & \textbf{0.768} & \textbf{0.065} & \textbf{5.949} & --    & --  &
& Ours                & 0.253          & --    & --  \\
\bottomrule
\end{tabular}%
}
\vspace{1pt}
\caption{\textbf{Quantitative comparison of image-to-material generation} (Left) and text-based material editing (Right) across datasets. User preference (\%) reports the fraction of comparisons where users preferred \textbf{Ours} over each baseline. GPT-5 and Users prefer our editing over baselines.}
\label{tab:comp_table}
\vspace{-3em}
\end{table}
\label{sec:baseline_comp}

We compare \ma\ against three state-of-the-art procedural material systems: BlenderAlchemy~\cite{huang2024blenderalchemy}, BlenderMCP~\cite{blendermcp}, and VLMaterial~\cite{livlmaterial}. 
We also report a Nearest-Neighbor baseline (the closest retrieved process) to verify that \ma\ isn't simply copying demonstrations. 
All baselines and \ma\ modules use a GPT-5 backbone for fairness.

For generation, perceptual similarity is measured following~\cite{livlmaterial} using CLIP cosine similarity, style loss, and sliced Wasserstein distance (SWD). 
For editing, where no reference render exists, we follow~\cite{huang2024blenderalchemy} and compute CLIP similarity between the edit instruction and the generated output. 
We additionally conduct GPT-5 and user forced-choice studies: for generation, evaluators receive the target and two outputs; for editing, they receive the initial material, the edit instruction, and two candidate results. 
We recruit 150 users on Amazon MT, with comparisons rated by three annotators and aggregated by majority vote.
\vspace{-0.7em}
\paragraph{Procedural Material Generation.}

\begin{figure*}[t]
\centering
\begin{minipage}{0.6\textwidth}
\centering
\includegraphics[width=\linewidth]{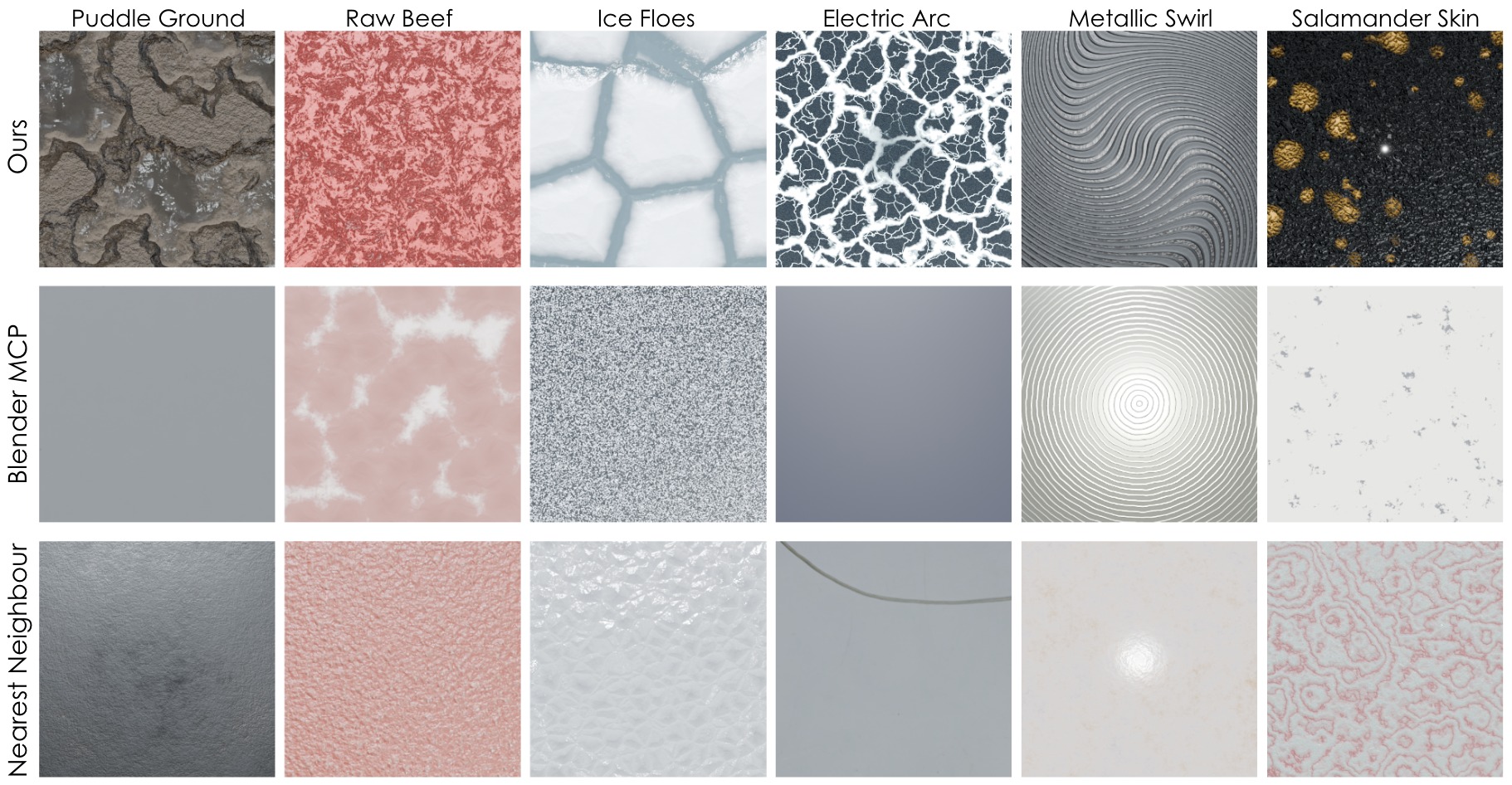}
\end{minipage}
\hfill
\begin{minipage}{0.35\textwidth}
\small
\vspace{12pt}
\caption{\small \textbf{Text-conditioned Material Generation.}
Our method produces visually diverse materials across a wide range of text prompts. 
Unlike BlenderMCP, our outputs align more closely with the user prompt while remaining distinct from the nearest retrieved example.
}
\label{fig:text-to-mat}
\end{minipage}
\vspace{-3em}
\end{figure*}

\begin{figure*}[t]
\centering

\begin{minipage}[t]{0.49\textwidth}
\centering
\includegraphics[height=2.6cm]{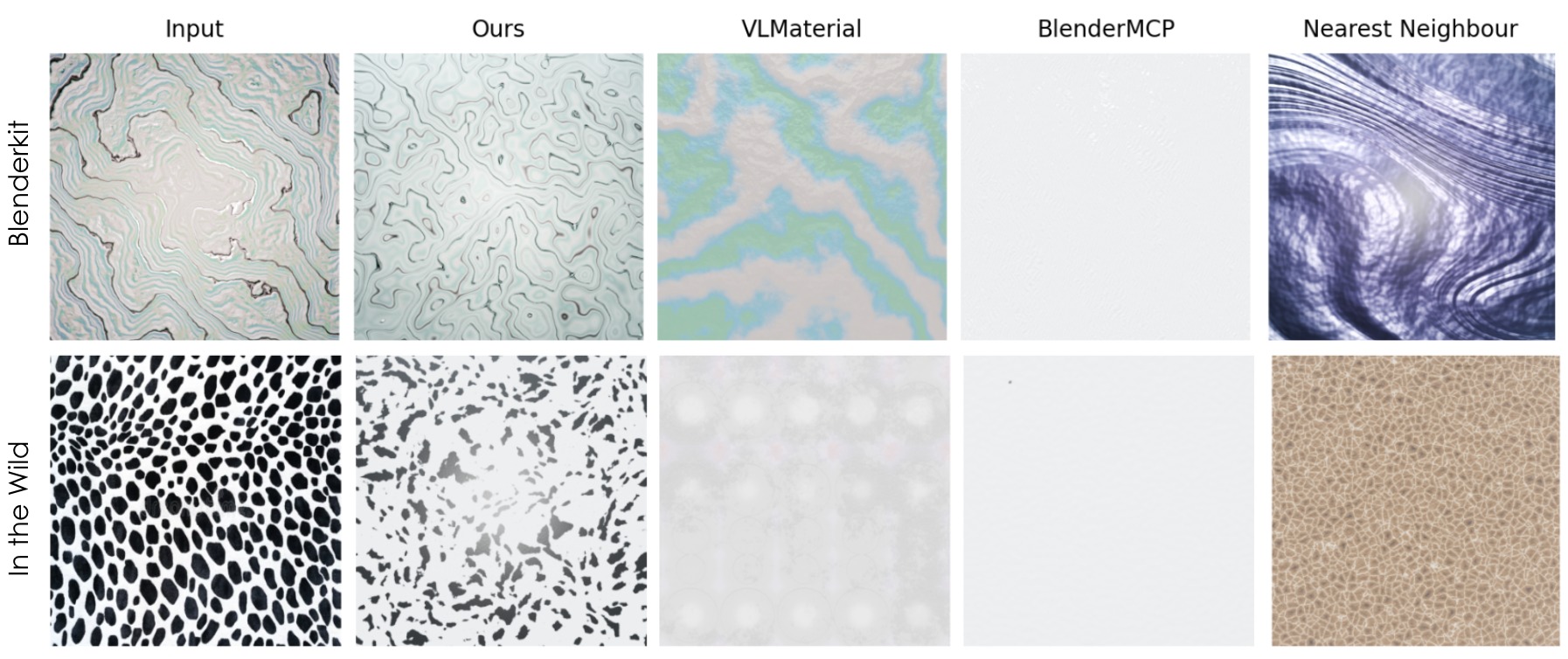}
\end{minipage}%
\hfill
\begin{minipage}[t]{0.49\textwidth}
\centering
\includegraphics[height=2.6cm]{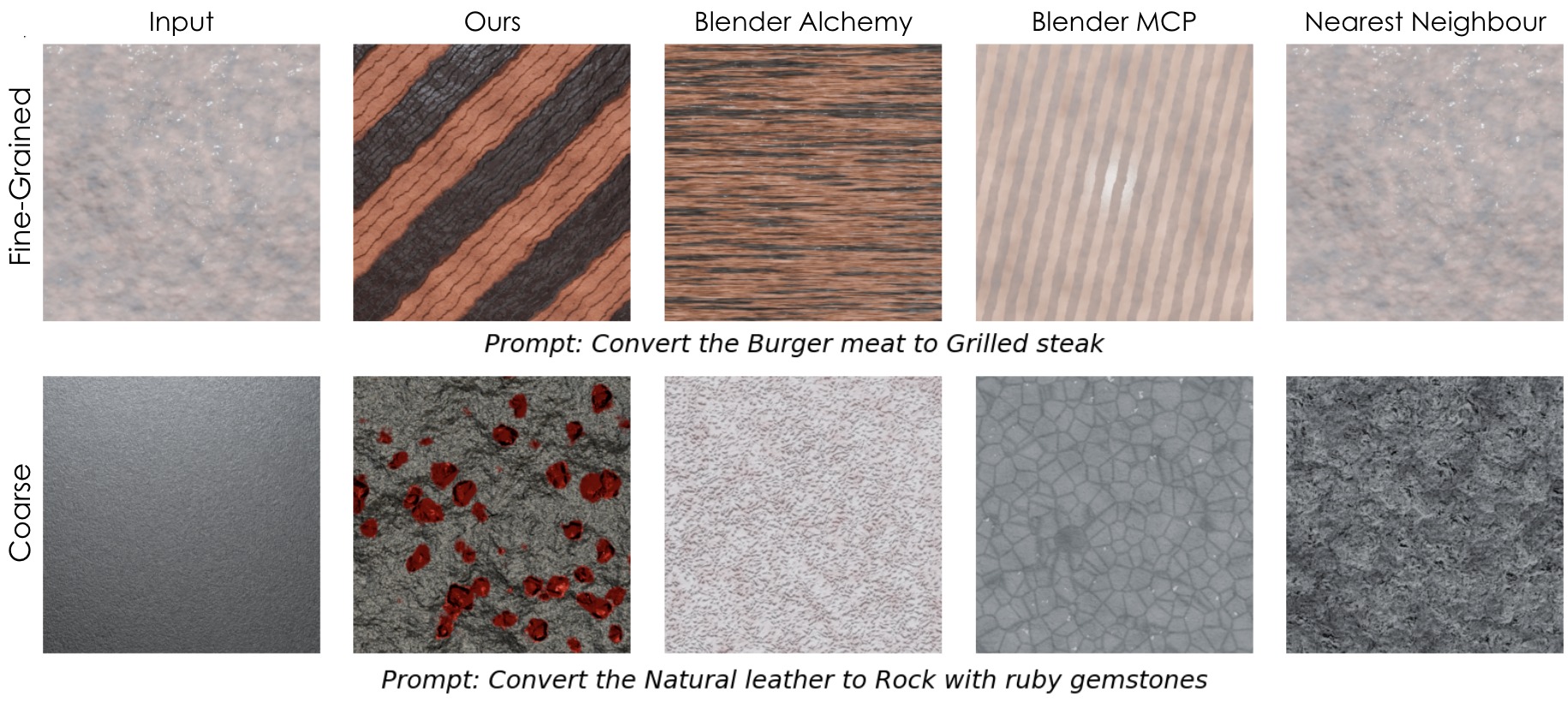}
\end{minipage}

\vspace{-0.3cm}




\caption{\small 
\textbf{Image conditioned Material Generation.} (Left) Images include nacre and dalmatian material. Our method better captures micro and macro details specified via the image reference.
\textbf{Text conditioned Material Editing.} (Right) Our method enables more precise and reliable manipulation of material graphs. For coarse edits, it makes the substantial structural changes required to achieve the intended transformation, while for fine edits, it can introduce subtle, localized modifications---producing nuanced effects that competing baselines fail to capture.
}
\label{fig:img_gen_txt_edit}
\vspace{-2em}
\end{figure*}

We evaluate all methods, excluding BlenderAlchemy (which does not support prompt-based generation), on two benchmarks: 
(i) 110 rendered materials from BlenderKit and 
(ii) 35 in-the-wild close-up photographs. 
Each prompt includes a short text description (e.g., “banana skin’’) and a reference image.
Figure~\ref{fig:text-to-mat} shows text-to-material generation results. 
\ma\ produces materials more closely aligned with textual prompts, achieving higher CLIP similarity (0.806 vs.\ 0.791 for nearest neighbor and 0.781 for BlenderMCP). Users prefer our results in 86\% and 92\% of comparisons against nearest neighbor and BlenderMCP, respectively. Table~\ref{tab:comp_table} reports quantitative results. 
\ma\ achieves the strongest performance on the challenging in-the-wild benchmark
and is competitive with VLMaterial on BlenderKit, where VLMaterial is fine-tuned.
CLIP, Style Loss, and SWD capture image-level similarity and are sensitive to scale,
alignment, and rendering, while GPT-5 and user preferences reflect prompt
faithfulness and procedural plausibility; these metrics therefore provide
complementary views of quality. Across both preference studies, \ma\ is
consistently favored over competing methods.

Figure~\ref{fig:img_gen_txt_edit} (left) shows qualitative comparisons. 
\ma\ reproduces both macro- and micro-scale structures—from nacre patterns to dalmatian textures—while VLMaterial often produces coarse approximations and BlenderMCP drifts from the target appearance. Our outputs remain clearly distinct from nearest-neighbor examples, indicating genuine synthesis rather than retrieval.
\vspace{-2em}

\paragraph{Procedural Material Editing.}
We evaluate BlenderAlchemy~\cite{huang2024blenderalchemy}, BlenderMCP~\cite{blendermcp}, and \ma\ on two editing tasks: 
25 coarse edits (e.g., wood $\rightarrow$ marble) and 
25 fine-grained edits (e.g., adding moss to brick), using base materials from BlenderKit. 
Coarse edits require substantial restructuring of the shader graph, while fine edits involve localized modifications.

As shown in Table~\ref{tab:comp_table}, GPT-5 consistently prefers edits produced by \ma\ over BlenderMCP and nearest-neighbor baselines. 
Qualitative examples (Fig.~\ref{fig:img_gen_txt_edit}, right) show that baselines often fail to incorporate key prompt attributes—for example, missing structural features—while \ma\ introduces both large-scale structural changes and subtle surface details.
\vspace{-2em}
\subsection{Ablations}
\label{sec:ablations}
\vspace{-0.5em}
We now analyze several aspects of \ma\, including its ability to generalize beyond the materials present in the process corpus and the effect of the number of retrieved demonstrations used during synthesis.


\begin{figure*}[t]
\centering
\includegraphics[width=\textwidth]{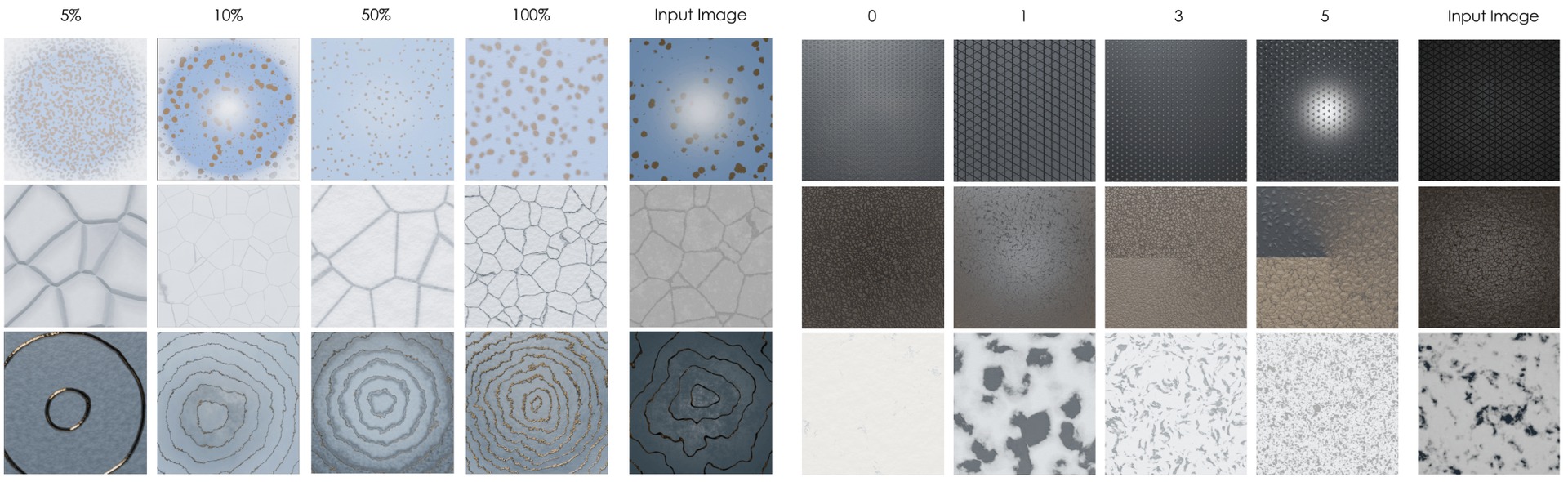}
\caption{\small 
\textbf{Effect of Video Corpus Size on Generation Quality.} (Left) Our retrieval-based approach maintains strong generation quality even when given only a small fraction of the tutorial corpus---evaluated at 5\% to 100\% availability. This demonstrates that \textsc{MaterialApprentice} can synthesize novel, high-quality materials by reflecting on only a handful of expert demonstrations.
\textbf{Effect of Number of Retrieved Demonstrations.} (Right) We evaluate \textsc{MaterialApprentice} using 0 (no retrieval), 1, 3, and 5 retrieved demonstrations. Quality improves from 0 to 3 retrieved process traces, with the largest gain from the first example; beyond three, improvements are marginal relative to the token cost.
}
\label{fig:ablation_figures}
\vspace{-1em}
\end{figure*}

\vspace{-0.5em}
\paragraph{Generalization beyond the process corpus.}
To test whether \ma\ relies on category-level cues from the corpus, we remove entire material categories corresponding to the evaluation prompts—Organic (blue jay egg), Concrete (cracked concrete), and Marble (midnight gold marble). Figure~\ref{fig:ablation_figures} shows that even under this setting, \ma\ continues to synthesize plausible procedural structures that capture the essential characteristics of the input images, indicating that the model generalizes procedural strategies rather than memorizing category-specific examples.
\vspace{-1em}
\paragraph{Effect of the number of retrieved demonstrations.}
We next vary the number of demonstrations provided as context during synthesis (Fig.~\ref{fig:ablation_figures}). Even a single retrieved process trace substantially improves generation quality compared to using no demonstrations. However, the benefit saturates quickly, with limited improvement beyond three retrieved examples, suggesting that a small number of relevant demonstrations is sufficient to guide effective process synthesis.
\vspace{-1em}

\begin{figure*}[t]
\centering
\begin{minipage}{0.35\textwidth}
\centering
\includegraphics[width=\linewidth]{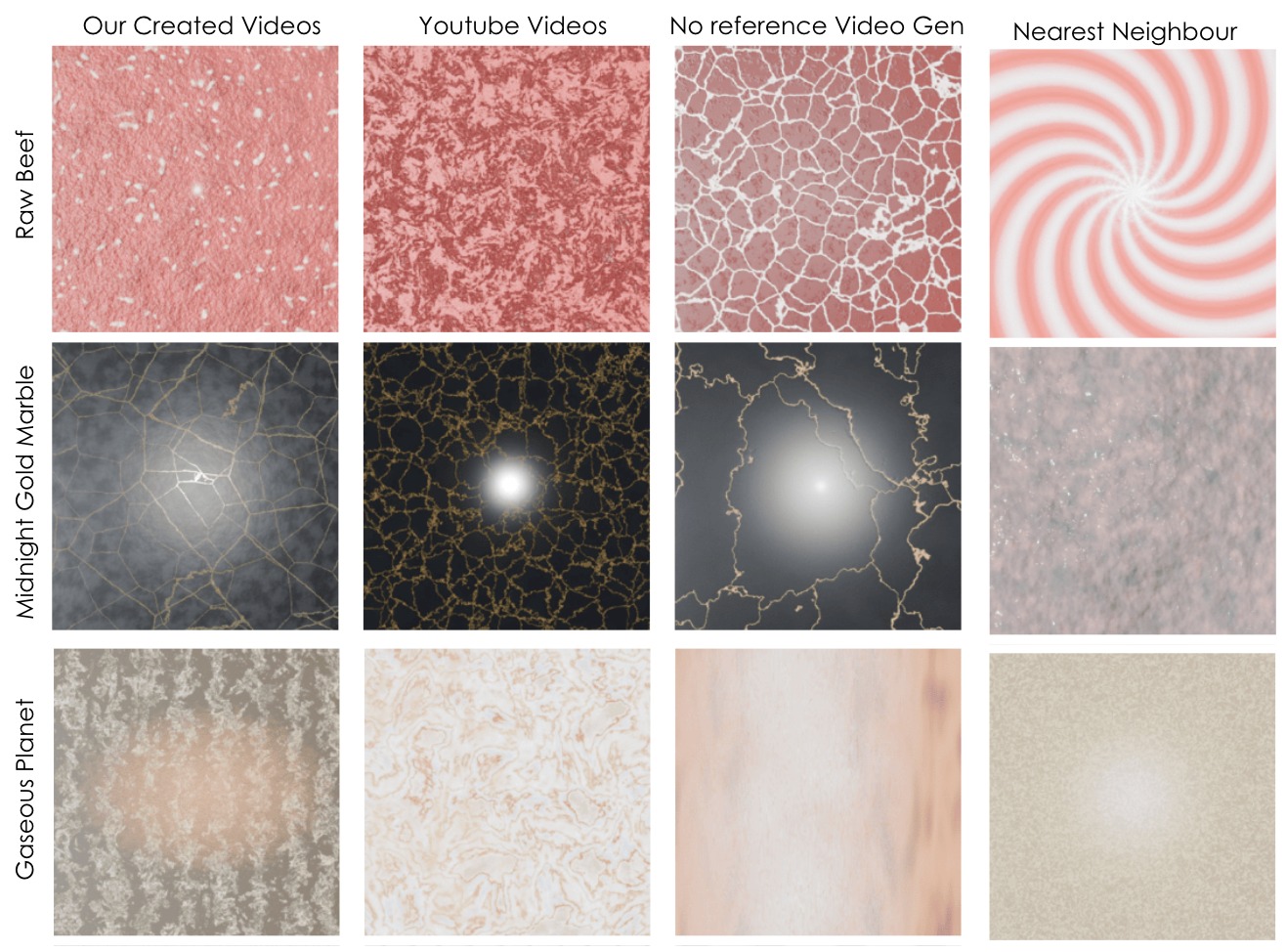}
\end{minipage}
\hfill
\begin{minipage}{0.6\textwidth}
\vspace{12pt}
\caption{\small \textbf{Generation using our curated video set.}
Materials generated by \ma\ using only our released tutorial videos as references (left column). Results remain strong across diverse prompts and differ clearly from YouTube-based generation, generation without references, and the nearest neighbor baseline.}
\label{fig:our_videos}
\end{minipage}
\vspace{-0.8cm}
\end{figure*}
\paragraph{Reproducibility with a curated video set.}
To further demonstrate reproducibility, we evaluate \ma\ using a curated set of tutorial videos that we created ourselves and release publicly with full rights. Figure~\ref{fig:our_videos} shows text-conditioned generation results when the reference pool is restricted to this small set. Despite the limited corpus, \ma\ continues to synthesize diverse materials that differ clearly from both nearest-neighbor retrieval and generation without demonstrations, indicating that even a handful of shareable tutorials provides sufficient process-level cues for effective material synthesis.
\vspace{-1em}

\begin{figure*}[t]
\centering
\includegraphics[width=\textwidth]{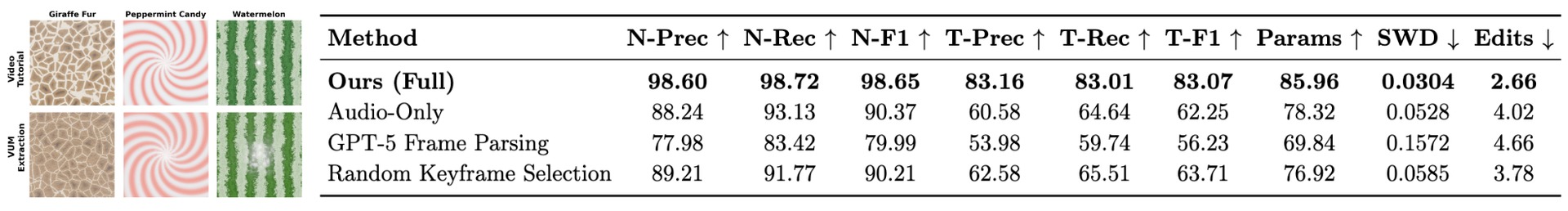}
\vspace{-0.6cm}
\caption{\small \textbf{Video analysis tools ablation.} 
(Left) Materials reconstructed from VAT-extracted process traces closely match those recreated by humans from the original tutorials. 
(Right) Quantitative comparison with three ablations: audio-only extraction, GPT-5 frame parsing, and random keyframe selection. 
Our full pipeline achieves the best node accuracy, topology reconstruction, parameter fidelity, perceptual similarity, and lowest edit distance.
}
\label{fig:vum_ablation_combined}
\vspace{-2.2em}
\end{figure*}

\paragraph{Impact of Video Analysis Tools.}
We analyze the contribution of our video analysis pipeline by constructing three variants: (1) \emph{Audio-only}, which ignores the visual stream and relies solely on narration; (2) \emph{GPT-5 Frame Parsing}, which replaces our specialized detectors with GPT-5-based frame interpretation; and (3) \emph{Random Keyframe Selection}, which removes our informativeness-based keyframe filtering. We evaluate all variants on 100 tutorial videos by reconstructing material graphs from extracted process traces and comparing them with ground-truth graphs. Node detection and topology reconstruction are evaluated using precision, recall, and F1 scores. Parameter fidelity is measured as the normalized absolute difference between predicted and ground-truth parameters, while perceptual similarity is computed using sliced Wasserstein distance (SWD). We additionally report graph edit distance, defined as the number of atomic operations required to transform the reconstructed graph into the ground-truth graph (node additions/removals, connection changes, and parameter adjustments), estimated by prompting an LLM to produce the minimal edit sequence. As shown in Fig.~\ref{fig:vum_ablation_combined}, our full pipeline consistently achieves the best performance across all metrics and produces graphs that most closely match the original materials. These results highlight the importance of jointly modeling visual structure, narration, and keyframe selection for reliable process extraction from tutorials.
\vspace{-1em}

\section{Conclusion}
\vspace{-0.5em}
Procedural material design is inherently process-driven: artists construct materials through layered operations, parameter tuning, and physical reasoning. We frame procedural material generation as retrieval-time process reasoning over expert demonstrations, elevating process to a first-class representation beyond graph-only synthesis. By retrieving process traces and grounding them into executable shader graphs, \ma\ produces materials that better reflect expert workflows while remaining interpretable and editable. More broadly, this perspective suggests generative systems that learn directly from creative practice rather than static artifacts alone. Extending process reasoning to multi-step design workflows, additional authoring tools beyond Blender, and larger corpora of demonstrations are promising directions for future work. Failure cases are discussed in the supplementary. Code, models, and data will be publicly released.

\section{Acknowledgments}
\vspace{-0.5em}
This work was supported in part by NSF grant 2402583 and gifts from Qualcomm, Google, and Adobe. We are grateful to the Blender experts who participated in our expert study: Barnabas Thomas, John Arquero, Benjamin Olawuni, Ricardo Raimundo, and Tanushree Roychoudhury. We also thank Krishna Chaitanya, Harsh Sinha, Kanav, and Diptimayee Gupta for their inspiration, support, and helpful discussions throughout this work.

\bibliographystyle{splncs04}
\bibliography{main}

\section{Supplementary Video}
Please watch the supplementary video for an overview of the method and highlights about key experiments. 

\section{Expert Study}
\subsection{Matched Graph-Space Baseline}
\label{sec:supp_graph_to_graph}

To further isolate the role of the intermediate representation, we evaluate a matched
\emph{Graph$\rightarrow$Graph} baseline in which both retrieval and generation operate
directly in graph space. This baseline retrieves material graphs and directly generates a material graph, removing the intermediate process representation. We compare it with \emph{Graph$\rightarrow$Process}, which retrieves graphs but synthesizes process traces, and our full \emph{Process$\rightarrow$Process} method, which retrieves and synthesizes in process space.

\paragraph{Evaluation protocol.}
We evaluate 17 materials across the three methods. For each material, artists are given the final generated graph from each method and perform two independent editing tasks. In \emph{Target Refinement}, artists edit the graph to better match the  reference material. In \emph{Instructional Edit}, artists apply a semantic edit using only parameter changes. Both tasks start from the same initial output graph and are limited to ten minutes. Separately, two artists choose the best result within each method triplet at three stages: the initial output, after target refinement, and after instructional edit.

\paragraph{Results.}
\begin{table}[t!]
\centering
\small
\setlength{\tabcolsep}{6pt}
\renewcommand{\arraystretch}{1.08}
\begin{tabular}{lccc}
\toprule
\textbf{Metric} & \textbf{G$\rightarrow$G} & \textbf{G$\rightarrow$P} & \textbf{P$\rightarrow$P} \\
\midrule
\multicolumn{4}{l}{\textbf{Preference (\%) $\uparrow$}} \\
Initial Output & 6 & 18 & \textbf{76} \\
After Target Refinement & 12 & 23 & \textbf{65} \\
After Instructional Edit & 18 & 0 & \textbf{82} \\
\midrule
\multicolumn{4}{l}{\textbf{Target-Refinement Effort $\downarrow$}} \\
Node edits (N) & 8.5 & 5.8 & \textbf{4.7} \\
Connection edits (C) & 23.5 & 19.9 & \textbf{15.3} \\
Parameter refinements (P) & \textbf{5.9} & 8.5 & 8.1 \\
Structural edits (N+C) & 32.0 & 25.7 & \textbf{20.0} \\
Total edits (N+C+P) & 37.9 & 34.2 & \textbf{28.1} \\
\midrule
\multicolumn{4}{l}{\textbf{Instructional-Edit Effort $\downarrow$}} \\
Total edits (parameters only) & \textbf{8.0} & 10.9 & 10.7 \\
\bottomrule
\end{tabular}
\caption{\small
Matched representation ablation comparing Graph$\rightarrow$Graph, Graph$\rightarrow$Process, and Process$\rightarrow$Process. Process$\rightarrow$Process is preferred at all stages and requires the fewest structural and total edits during target refinement. Graph$\rightarrow$Graph requires fewer parameter-only instructional edits, but its final edited outputs are rarely preferred, suggesting that fewer exposed controls do not imply better editability.
}
\label{tab:graph_to_graph_ablation}
\end{table}
Table~\ref{tab:graph_to_graph_ablation} shows that \emph{Process$\rightarrow$Process}
is preferred at all stages: 76\% for the initial output, 65\% after target refinement,
and 82\% after instructional edit. It also requires the fewest node edits, connection
edits, structural edits, and total edits during target refinement, indicating that
process-space reasoning produces graphs that are closer to the intended procedural
structure. \emph{Graph$\rightarrow$Graph} requires fewer parameter-only edits during
instructional editing, but this reflects the fact that its simpler graphs expose fewer
tunable controls; despite requiring fewer parameter tweaks, its final edited outputs are
rarely preferred. These results show that the advantage of process-space reasoning
persists under a matched graph-space baseline and improves both initial graph quality
and artist-facing editability.

\subsection{Parameter-only Semantic Editing}
\label{sec:supp_parameter_editing}

To further evaluate whether generated materials expose useful controls, we conduct
a parameter-only semantic editing study. Unlike the main expert study, where
artists may modify graph structure, this study fixes the generated material graph
and permits only parameter changes. This isolates whether the graph produced by
\ma\ contains editable controls that support meaningful downstream refinement.

We evaluate 6 generated materials with 4 semantic edits per material, for a total
of 24 editing tasks. Two Blender artists perform each edit using only exposed
parameters, with at most 3 refinement attempts and a 5-minute time limit per edit.
After each task, artists rate satisfaction with the final edit and ease of use on a
5-point Likert scale.

Artists report an average satisfaction score of 4.10/5 and an ease-of-use score of
3.75/5, with an average editing time of 1:48 minutes per edit. These results suggest
that \ma\ does not merely produce executable graphs, but also exposes parameters
that support efficient semantic refinement without requiring structural graph edits.

\subsection{Qualitative Examples}
\begin{figure*}[t!!]
    \centering
    \begin{overpic}[width=0.98\textwidth]{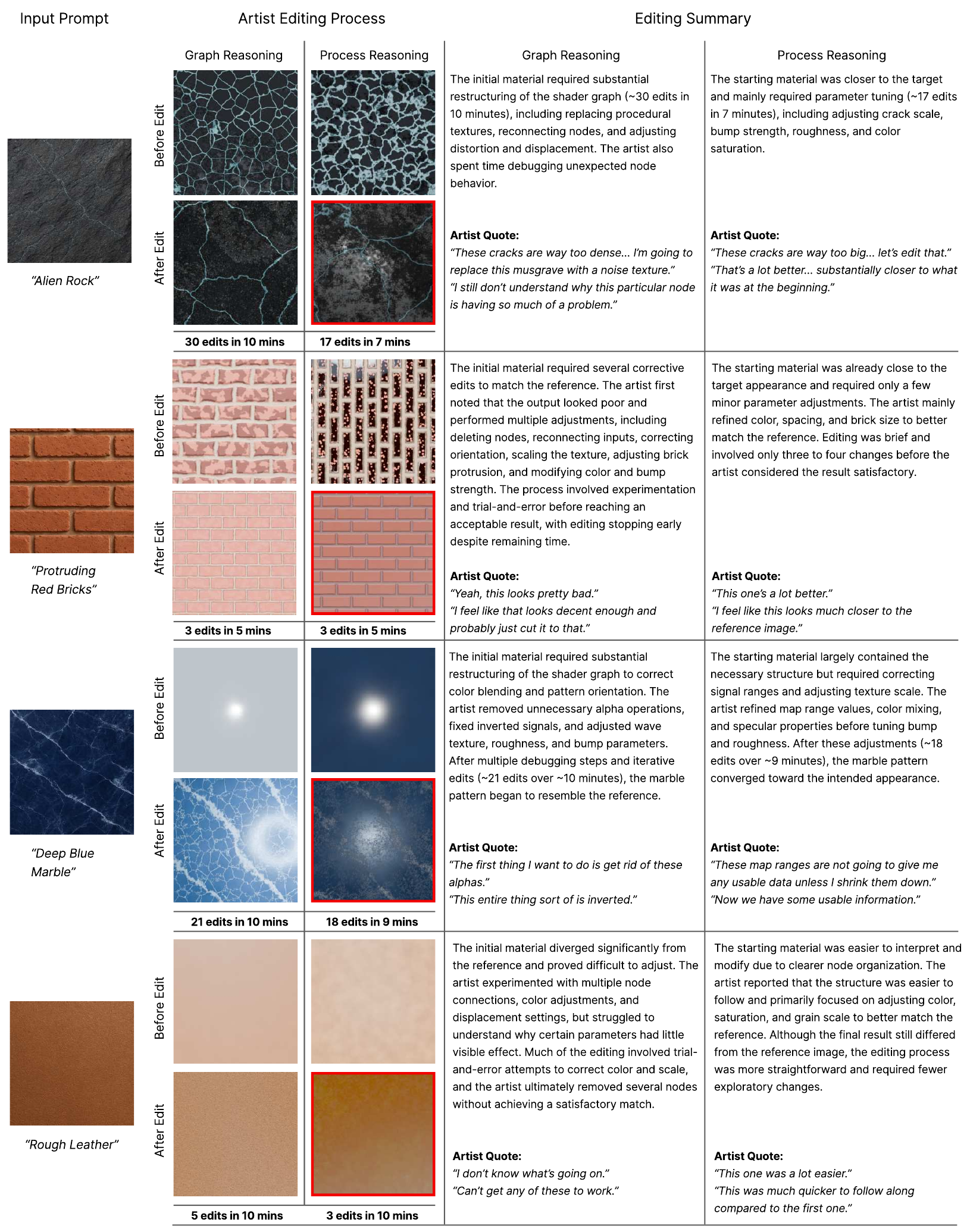}
    \end{overpic}
    \caption{\small Expert study comparing graph retrieval and process retrieval. Artists first constructed material graphs by following the generated process traces and then edited them to better match the target prompt. Edit count and editing time serve as proxies for process quality. Across examples, process-based retrieval produces materials that start closer to the target, require fewer edits, and are consistently preferred by artists (red borders), while graph retrieval often requires substantial restructuring of the shader graph. Artists also reported that process-derived graphs were easier to interpret and modify.}
    \vspace{-0.3cm}
    \label{fig:expert_study_qualitative}
    \vspace{-0.1cm}
\end{figure*}

\label{sec:supp_expert_qual}

Figure~\ref{fig:expert_study_qualitative} provides representative examples from the expert study comparing graph-conditioned and process-conditioned retrieval. For each prompt, artists first constructed a material graph by following the generated process trace and then refined the graph to better match the target material. The figure shows the input prompt, the initial and edited materials, the number of edits and editing time, and summarized think-aloud feedback from the artists.

Across these examples, process-conditioned retrieval typically produces graphs that start closer to the target appearance and require fewer corrective edits. For instance, in the \emph{Alien Rock} and \emph{Rough Leather} examples, graph-conditioned retrieval leads to substantial restructuring, including replacing procedural textures, reconnecting nodes, and debugging unexpected node behavior. In contrast, process-conditioned retrieval more often yields a graph with the right high-level structure, allowing artists to focus on parameter tuning such as scale, color, roughness, and bump strength.

The artist comments further illustrate this difference. Graph-conditioned outputs often cause confusion or trial-and-error editing when the node organization is difficult to interpret or when parameters have unclear effects. Process-conditioned outputs are described as easier to follow and quicker to refine, suggesting that process traces transfer not only node-level operations but also higher-level construction logic. These qualitative examples complement the quantitative results in Table~\ref{tab:expert_workflow_analysis}, showing how process-conditioned retrieval reduces structural repair and improves artist-facing editability.

\section{Video Tutorial Corpus}
\begin{table}[t]
\centering
\begin{minipage}{0.60\linewidth}
\centering
\tiny
\renewcommand{\arraystretch}{0.9}
\setlength{\tabcolsep}{3pt}
\begin{tabular}{l|ccc}
\hline
\textbf{Category} & \textbf{\#} & \textbf{Avg Nodes} & \textbf{Video (min)} \\
\hline
Animal & 5 & 15.8 & 10.7 \\
Asphalt & 2 & 22.0 & 19.0 \\
Bricks & 5 & 13.4 & 12.4 \\
Ceramic & 2 & 10.5 & 8.0 \\
Concrete & 4 & 19.5 & 13.5 \\
Fabric & 10 & 10.8 & 9.7 \\
Floor & 3 & 20.0 & 20.7 \\
Food & 11 & 11.5 & 11.8 \\
Fruit & 9 & 15.6 & 20.9 \\
Fx & 8 & 14.1 & 10.5 \\
Glass & 6 & 9.8 & 9.2 \\
Grass & 1 & 7.0 & 8.0 \\
Ground & 5 & 17.6 & 15.0 \\
Human & 4 & 14.3 & 11.8 \\
Ice & 4 & 13.0 & 14.5 \\
Leather & 3 & 15.7 & 11.3 \\
Liquid & 5 & 11.8 & 8.4 \\
Marble & 5 & 16.6 & 15.0 \\
Metal & 14 & 15.4 & 10.1 \\
Organic & 5 & 14.0 & 9.6 \\
Ornaments & 9 & 13.3 & 8.7 \\
Paper & 2 & 11.0 & 12.0 \\
Paving & 4 & 17.3 & 16.3 \\
Plaster & 4 & 13.0 & 12.5 \\
Plastic & 9 & 15.8 & 11.1 \\
Rock & 8 & 15.6 & 13.0 \\
Sand & 4 & 16.3 & 12.5 \\
Tech & 4 & 11.3 & 15.5 \\
Wood & 3 & 13.3 & 12.7 \\
\hline
\end{tabular}
\end{minipage}
\hfill
\begin{minipage}{0.35\linewidth}
\caption{Statistics of the tutorial video dataset used by \ma\ to capture tacit expertise during procedural material generation.}
\label{tab:yt-dataset}
\end{minipage}
\end{table}

\begin{table}[t]
\centering
\begin{minipage}{0.58\linewidth}
\centering
\tiny
\renewcommand{\arraystretch}{0.9}
\setlength{\tabcolsep}{3pt}
\begin{tabular}{l|c|c}
\hline
\textbf{Category} & \textbf{BlenderKit} & \textbf{In-the-Wild} \\
\hline
Animal & 6 & 8 \\
Asphalt & 2 & 1 \\
Bricks & 5 & 0 \\
Ceramic & 3 & 1 \\
Concrete & 4 & 1 \\
Fabric & 5 & 3 \\
Floor & 5 & 0 \\
Food & 8 & 0 \\
Fruit & 6 & 3 \\
Fx & 9 & 1 \\
Glass & 4 & 0 \\
Grass & 1 & 0 \\
Ground & 4 & 0 \\
Human & 1 & 1 \\
Ice & 3 & 0 \\
Leather & 5 & 1 \\
Liquid & 4 & 0 \\
Marble & 7 & 0 \\
Metal & 8 & 4 \\
Organic & 2 & 1 \\
Ornaments & 4 & 2 \\
Plastic & 6 & 1 \\
Plaster & 2 & 0 \\
Rock & 5 & 1 \\
Sand & 1 & 0 \\
Tech & 3 & 0 \\
Wood & 0 & 5 \\
Paving & 0 & 1 \\
\hline
\end{tabular}
\end{minipage}
\hfill
\begin{minipage}{0.38\linewidth}
\caption{Category distribution of test prompts across the BlenderKit and In-the-Wild datasets used in our evaluation. The table reports the number of materials per category in each dataset.}
\label{tab:text_prompts_stats}
\end{minipage}
\end{table}
To enable \ma\ to reflect on expert workflows, we curate a tutorial video corpus containing \textbf{158} publicly available Blender material–creation tutorials from established YouTube creators such as Ryan King Art, Ducky 3D, Sam Bowman, PIXXO 3D, BlenderBiteSize, and others. These videos provide diverse demonstrations of real expert practice that \ma\ retrieves at inference time.

The corpus spans 29 material categories, with each category averaging 5.5 tutorials. Tutorials have an average duration of 12.57 minutes, and the demonstrated materials contain on average 14.32 nodes, offering rich signals about procedural structure, parameter tuning, and tacit artistic reasoning (see Table~\ref{tab:yt-dataset}). We will release the YouTube IDs and category labels for all videos, allowing the corpus to be reconstructed while preserving attribution to the original creators.

This corpus is strictly used for retrieving process traces at inference time. Experiments in section 4.3/ Figure 10 (main paper) show that \ma\ is able to generate novel materials with even a small fraction of the corpus which is further demonstrated by  running \ma\ on the corpus of videos made by us (Figure 9).

\section{Video Analysis Tools for Process Extraction}
\label{sec:supp_video_tools}

This section provides implementation details for the video analysis pipeline used to extract process traces from tutorial videos. As discussed in Sec.~3.4 of the main paper, this component is primarily an engineering step that converts demonstrations into structured signals usable for retrieval-time process reasoning.

Blender tutorial videos exhibit a consistent box-and-wire interface, allowing procedural structure to be reliably recovered using standard computer vision tools. Our pipeline automatically decomposes each tutorial video $v$ into frames and narration, reconstructs intermediate node graphs, and summarizes these multimodal signals into a textual process trace $p_v$ describing the material construction workflow.

The pipeline relies on lightweight pretrained detectors and segmentation models fine-tuned on Blender shader editor screenshots. Because the interface exhibits highly consistent layout, color schemes, and typography, standard architectures achieve high accuracy without extensive tuning. All stages operate automatically and require no manual annotation. We note that minor perception errors rarely propagate to the final generation stage, as the \ps\ reasons over construction patterns rather than exact node configurations.

\subsection{Video Decomposition}

Each tutorial video $v$ is decomposed into visual frames and audio narration. Let 
$X_v = \{(t_k, I_k)\}_{k=1}^{T_v}$ denote the frame sequence and $a_v$ the audio stream.

\paragraph{Keyframe Selection.}

\begin{figure}
    \centering
    \includegraphics[width=\linewidth]{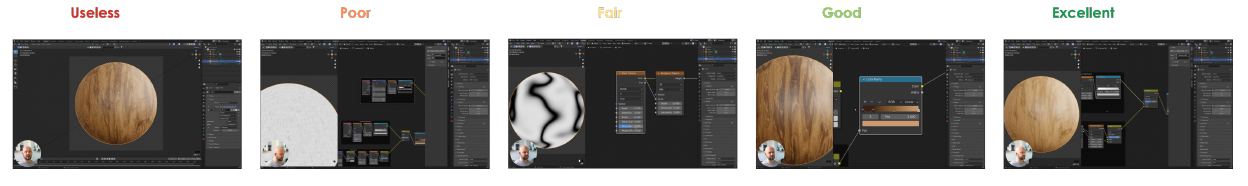}
    \caption{\small Illustration of frame informativeness used for keyframe selection. Frames range from \textit{useless} to \textit{excellent} depending on their suitability for node-graph reconstruction. Low-quality frames exhibit occlusion, extreme zoom, or limited graph visibility, whereas high-quality frames clearly expose node layouts, connections, and parameter controls required for parsing.}
    \label{fig:score_samples}
\end{figure}

Tutorial videos often contain frames that are difficult to parse due to occlusions, extreme zoom, or limited visibility of the node graph. We therefore first identify frames that are suitable for reliable graph reconstruction. Frames are categorized according to their suitability for parsing, ranging from \textit{useless} to \textit{excellent}, as illustrated in Fig.~\ref{fig:score_samples}.

\begin{figure}
    \centering
    \includegraphics[width=\linewidth]{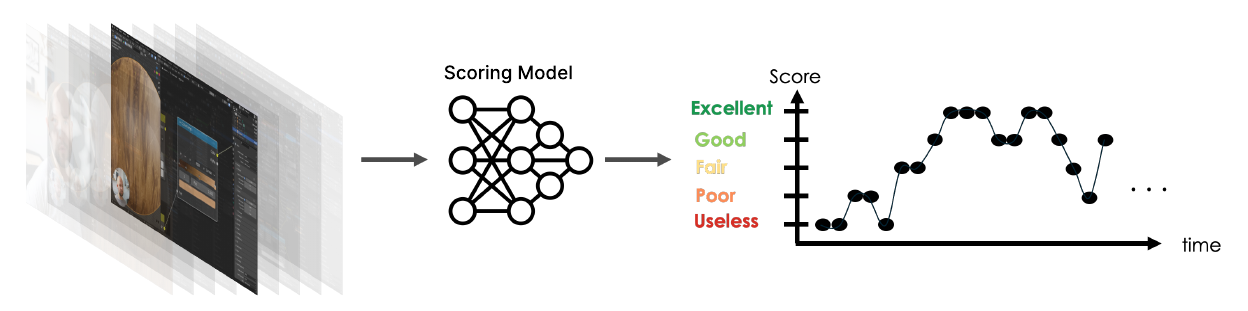}
    \caption{\small Keyframe scoring pipeline. Video frames are evaluated by a scoring model that assigns an informativeness score based on node visibility, parameter readability, and graph coverage. The resulting score sequence over time identifies frames suitable for reliable graph reconstruction.}
    \label{fig:score_model}
\end{figure}

To automatically estimate this quality, each frame $I_k$ is assigned an informativeness score
\begin{equation}
s_k = \Phi_{\text{score}}(I_k),
\end{equation}
which evaluates node visibility, text readability, graph coverage, and overall visual clarity. Technically, this is achieved via the scoring network $\Phi_{\text{score}}$ uses a YOLOv8n~\cite{yolov8} backbone fine-tuned as a five-class classifier to predict frame informativeness (\textit{excellent}, \textit{good}, \textit{fair}, \textit{poor}, \textit{useless}). The scoring process is illustrated in Fig.~\ref{fig:score_model}. We train the score model on our internally created and annotated dataset consisting of $1000$ synthesized Blender screenshots labeled across the five categories for 50 epochs at $1080$p resolution with a batch size of 128, achieving $70\%$ top-1 accuracy. 

Frames with scores above a threshold $\tau_s$ form a candidate set
\begin{equation}
\mathcal{C}_v = \{(t_k,I_k)\mid s_k \ge \tau_s\}.
\end{equation}
Even after filtering, tutorials may still contain many visually similar frames where the node graph remains unchanged while the instructor explains a concept. To reduce redundancy and improve efficiency, candidate frames are embedded using $\Phi_{\text{emb}}$ (CLIP) and clustered by cosine similarity. The final frame from each cluster is retained, capturing the most evolved graph state at that stage of the tutorial.

\paragraph{Narration Transcription.}

The audio stream $a_v$ is transcribed using the Whisper-Large automatic speech recognition model, producing a time-aligned transcript
\begin{equation}
\tau_v = \Phi_{\text{asr}}(a_v).
\end{equation}
The outputs of this stage are the keyframe set $X_v^{\mathcal{K}}$ and the aligned narration transcript $\tau_v$.

\subsection{Graph Reconstruction}

For each selected keyframe $I_k$, the pipeline reconstructs the corresponding node-graph state by detecting nodes, recovering UI parameters, and inferring graph connectivity.

\paragraph{Image-Level Parsing.}

\begin{wrapfigure}{r}{0.55\linewidth}
\vspace{-10pt}
\centering
\includegraphics[width=\linewidth]{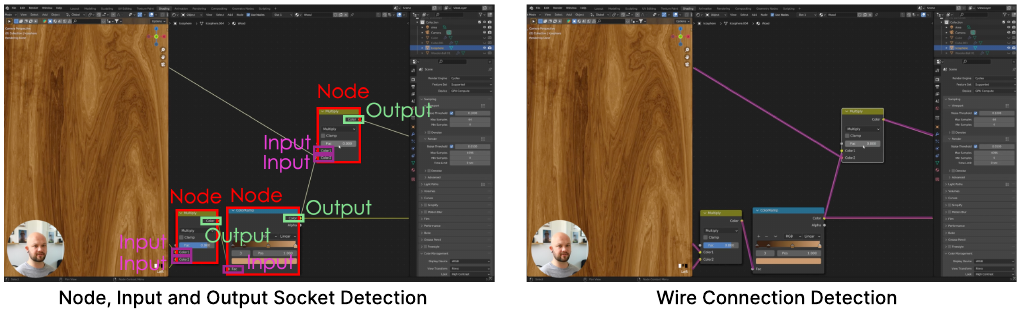}
\label{fig:image_detection}
\vspace{-10pt}
\end{wrapfigure}

We first identify the structural components of the shader graph. An object detector localizes nodes together with their input and output sockets, while a segmentation model extracts connection wires (see figure on the right):

\begin{equation}
\mathcal{N}_k,\mathcal{S}_k^{in},\mathcal{S}_k^{out} =
\Phi_{\text{det}}^{img}(I_k),
\quad
\mathcal{W}_k = \Phi_{\text{seg}}^{wire}(I_k).
\end{equation}

OCR is then applied within detected node regions to recover node names and socket labels.

\paragraph{Node-Level Parameter Extraction.}

\begin{wrapfigure}{l}{0.48\linewidth}
\vspace{-10pt}
\centering
\includegraphics[width=\linewidth]{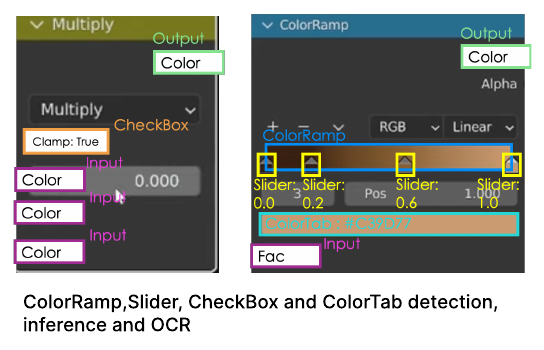}
\label{fig:node_detection}
\vspace{-10pt}
\end{wrapfigure}

Each detected node region $I_n$ is further analyzed to recover UI elements encoding parameter values. A lightweight node-level detector identifies interface components such as sliders, color ramps, checkboxes, and color tabs (see figure on the left):

\begin{equation}
\mathcal{U}_n = \Phi_{\text{det}}^{node}(I_n).
\end{equation}

Slider positions are normalized within their UI bounds, while color tabs are converted to RGB values, producing node parameter states $\Theta_n$.

Both the image-level detector $\Phi_{\text{det}}^{img}$ and the node-level detector $\Phi_{\text{det}}^{node}$ use YOLOv8n trained at $1080$p resolution for 50 epochs. Both achieve \textit{mAP$_{50}$} scores near $0.98$ with precision and recall in the $0.95$–$0.98$ range. Connection wires are segmented using Mask2Former~\cite{mask2former} (Swin-B backbone) trained on $950/50$ train/validation frames using AdamW with cosine decay and standard augmentations, achieving a mean IoU of $0.87$.

\paragraph{Topology Reconstruction.}

\begin{wrapfigure}{r}{0.45\linewidth}
\vspace{-30pt}
\centering
\includegraphics[width=\linewidth]{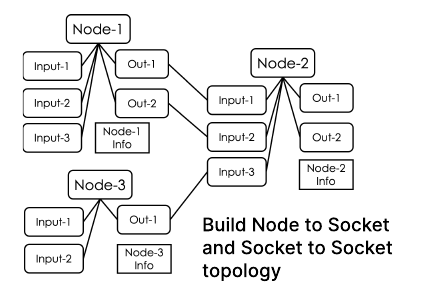}
\label{fig:graph_building}
\vspace{-10pt}
\end{wrapfigure}

Graph connectivity is inferred from geometric relationships between nodes and sockets. Socket–node associations are determined via spatial overlap, while socket–socket connections are inferred by fitting candidate Bézier curves between output and input sockets and measuring their overlap with the wire segmentation mask.

Combining detected nodes, sockets, inferred edges, and parameter states yields the graph representation (see figure on the right):

\begin{equation}
G_k = (\mathcal{N}_k, \mathcal{S}_k, E_k, \Theta_k),
\end{equation}

where $E_k$ represents graph connectivity and $\Theta_k$ stores node parameters. Each pair $(t_k,G_k)$ therefore represents a temporally grounded snapshot of the evolving material graph.

\subsection{Process Summarization}

Finally, the sequence of reconstructed graph states is fused with the narration transcript to produce a textual process trace. Given $\{(t_k,G_k)\}$ and the narration transcript $\tau_v$, a language-model summarizer integrates these multimodal signals:

\begin{equation}
p_v = \Phi_{\text{sum}}(\{(t_k,G_k)\}, \tau_v).
\end{equation}

The summarizer aligns narrated explanations with observed graph edits and produces a compact description of the construction process. Self-refinement prompts ensure that the generated process remains consistent with both the transcript and the reconstructed graph sequence. The resulting trace $p_v$ captures both explicit construction steps and the procedural reasoning expressed by the artist.

Figures~\ref{fig:transcript} and~\ref{fig:procedure} show example outputs before and after summarization. Raw narration transcripts are typically long (often exceeding $60$k tokens for a $10$ minute tutorial), while the summarized process traces are significantly shorter (typically $\sim8$k tokens), making them easier to condition the \ps\ during retrieval-time reasoning.

\section{Test Prompts used for Evaluation}

All experimental evaluations—including text-conditioned generation, image-conditioned generation, and ablation studies—use materials sourced from the BlenderKit dataset. Table~\ref{tab:text_prompts_stats} summarizes the distribution of test prompts across the 29 BlenderKit material categories used in our benchmark, averaging 3.9 prompts per category. The source materials used in both coarse and fine-grained editing experiments are also drawn from BlenderKit.

For image-conditioned generation, we additionally evaluate on a set of in-the-wild close-up material photographs collected from the internet. The category distribution of these images is also reported in Table~\ref{tab:text_prompts_stats}.

\section{Ablations}

In addition to the ablations presented in the main paper, we analyze the contribution of the video analysis tools used for process extraction, the effect of summarizer self-refinement, and the role of the \dcom\ in enabling reliable editing.

\subsection{Ablation of Video Analysis Tools}
The video analysis pipeline converts a tutorial video into a structured process trace through three stages. 
First, \textbf{Video Decomposition} extracts frames and audio, scores frames for informativeness, removes near-duplicates using perceptual embeddings, and produces a time-aligned transcript via ASR. 
Second, \textbf{Graph Reconstruction} parses each selected keyframe using lightweight detectors, segmentors, and OCR to recover the evolving node graph—its nodes, socket connections, and parameter states such as sliders and color ramps. 
Finally, \textbf{Process Summarization} fuses these graph states with the narration using an LLM with self-refinement to produce a concise textual process that captures both explicit editing steps and the expert’s underlying rationale.

To assess the role of each stage, we evaluate four ablations of this pipeline and analyze their impact on text-conditioned material generation:

\begin{itemize}
\item \textbf{Audio-only}: removes all visual analysis and reduces the pipeline to ASR on the narration followed by process summarization.
\item \textbf{Random Keyframe Sampling}: replaces the scored and deduplicated keyframe selection in video decomposition with uniform random sampling.
\item \textbf{GPT-based Frame Parsing}: replaces the domain-adapted detectors and segmentors used for graph reconstruction with GPT-5-based vision parsing of each frame.
\item \textbf{w/o Summarizer Self-Refine}: removes the iterative self-refinement stage in process summarization, forcing the summarizer to compress the raw multimodal trace in a single pass.
\end{itemize}

Each ablation isolates a single stage while keeping the rest of the pipeline unchanged. The resulting process traces are then used by \ma\ during material generation.

\begin{figure}[t!]
    \centering
    \begin{overpic}[width=\textwidth]{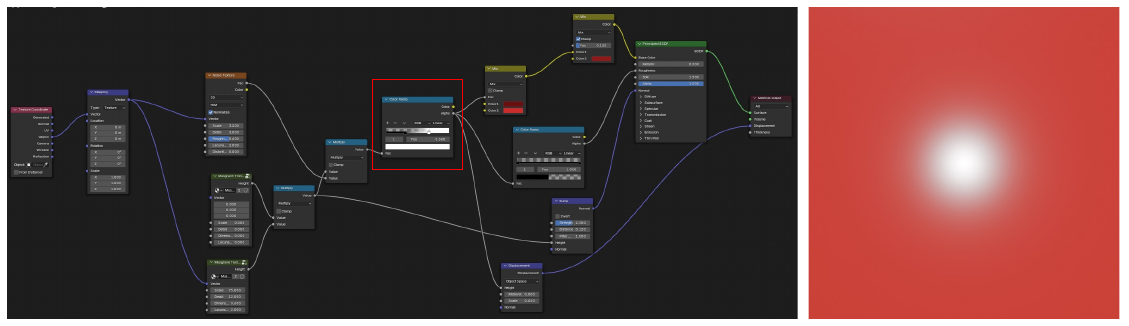}
    \end{overpic}
    \caption{\small\textbf{Audio-only ablation.} Without visual parsing, the system relies solely on spoken narration to reconstruct the process, producing an oversimplified and structurally incorrect node graph (left). Lacking spatial and topological cues, multiple semantic roles—color, roughness, bump, and displacement—collapse into a single reused mask, resulting in a flat, low-fidelity material (right). This illustrates that narration alone is insufficient for recovering procedural workflows, as critical graph structure is conveyed visually rather than verbally.}
    \label{fig:VUMablation_audio}
\end{figure}

\begin{figure}[h]
    \centering
    \begin{overpic}[width=\textwidth]{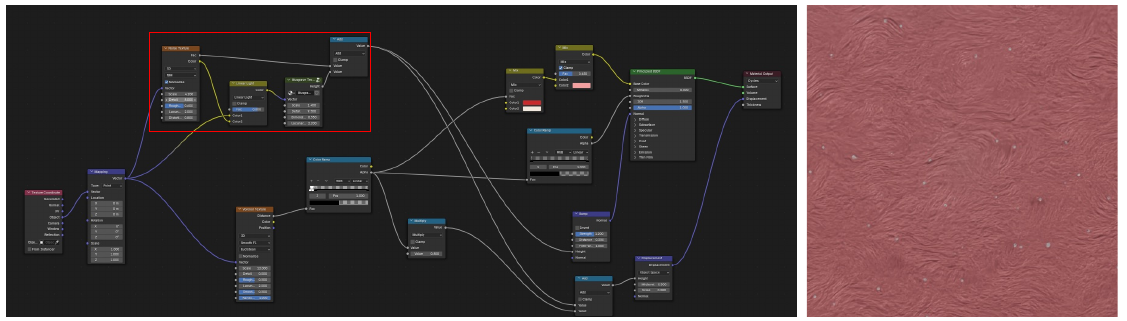}
    \end{overpic}
    \caption{\small\textbf{Random keyframe sampling ablation.} Without importance-weighted keyframe selection, the pipeline receives randomly sampled frames that often miss critical graph-level relationships. While many correct primitives (e.g., Noise, Musgrave, Voronoi) are detected, they are assembled with incorrect topology, producing a diffuse, isotropic texture rather than the elongated fiber structure of raw beef. This highlights the importance of selecting informative keyframes for recovering coherent procedural structure.}
    \label{fig:VUMablation_randomsample}
\end{figure}

\begin{figure}[h]
    \centering
    \begin{overpic}[width=\textwidth]{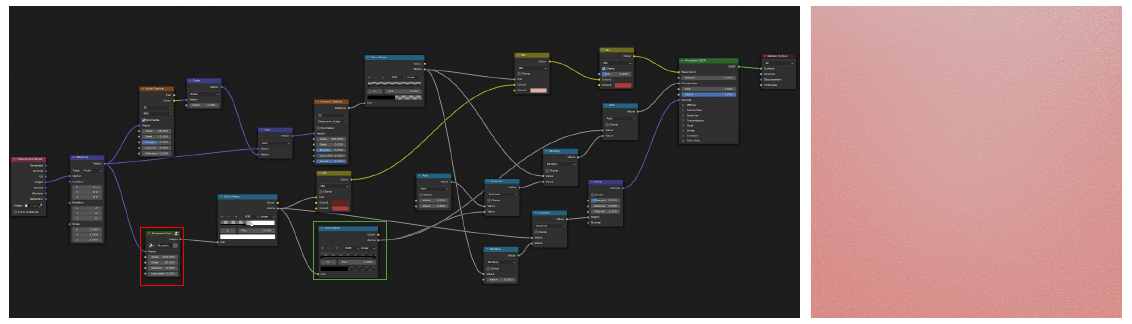}
    \end{overpic}
    \caption{\small\textbf{GPT-5 Frame Parsing Ablation :} GPT-5–based frame parsing introduces systematic parameter and color-value misinterpretations. Although the model identifies high-level node types, it frequently misreads slider values, color-ramp stops, and vector–color outputs, leading to incorrect frequency scales and collapsed shading cues. The resulting graph (left) contains plausible structure but incorrect numeric detail, producing an over-sharpened, sponge-like material (right) rather than the intended raw-beef appearance.
    }
    \label{fig:VUMablation_gpt}
\end{figure}

\begin{figure}[h]
    \centering
    \begin{overpic}[width=\textwidth]{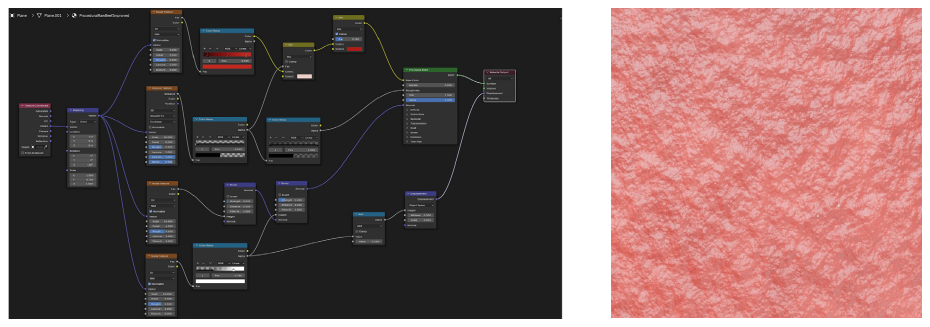}
    \end{overpic}
    \caption{\small\textbf{No summarizer self-refinement ablation.} 
Without the self-refinement loop, the summarizer compresses the multimodal trace too aggressively, dropping key procedural cues and weakening dependencies between graph components. Although most nodes and masks are correctly identified, their functional relationships are reconstructed imprecisely. The resulting material resembles ground or processed meat rather than structured raw beef, illustrating that iterative refinement is important for preserving long-range dependencies across color, roughness, bump, and displacement.}
    \label{fig:VUMablation_norefine}
\end{figure}

\begin{figure}[h]
    \centering
    \begin{overpic}[width=\textwidth]{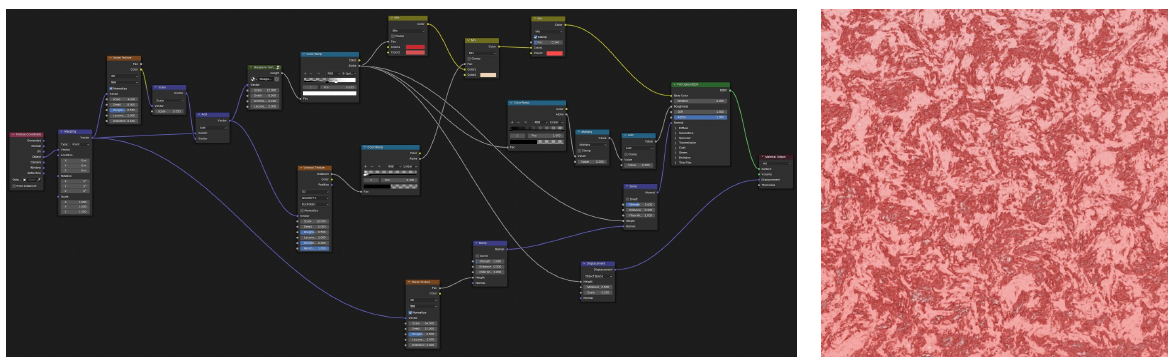}
    \end{overpic}
    \caption{\small\textbf{Full pipeline.} 
Combining importance-weighted keyframe selection, domain-adapted frame parsing, accurate parameter extraction, and iterative summarization enables faithful reconstruction of the procedural logic in the tutorial. The recovered graph captures the intended multi-scale structure—global strand flow, mid-frequency muscle fibers, Voronoi fat marbling, and micro-scale surface grain—with consistent modulation across color, roughness, bump, and displacement. The resulting render exhibits anatomically plausible raw-beef structure, showing that all components of the extraction pipeline are necessary to recover expert procedural workflows.}
    \label{fig:VUMablation_full}
\end{figure}

Figures~\ref{fig:VUMablation_audio}--\ref{fig:VUMablation_full} illustrates how these components affect generation quality. 
\textbf{Audio-only} performs the worst: without spatial information about the node graph, the model cannot resolve references such as “connect the output of the color ramp to the mix factor,” especially when multiple such nodes exist, resulting in severely degenerate graphs. 
\textbf{Random Keyframe Sampling} improves over \textbf{Audio-only} by exposing the correct set of nodes, but randomly selected frames rarely capture meaningful structural cues—many contain isolated nodes, occlusions, or uninformative UI states—leading to incorrect or incomplete graph connectivity. 
\textbf{GPT-based Frame Parsing} further improves quality since it analyzes visual frames rather than relying purely on narration, but generic vision parsing often misreads Blender UI elements such as ColorRamp stops, color tabs, and slider values, producing inaccurate graph states that degrade the extracted procedural cues. 
\textbf{w/o Summarizer Self-Refine} starts from the strongest multimodal trace, but a single-pass summary often overlooks subtle design rationale and parameter intent, yielding weaker guidance for generation.
In contrast, the full pipeline—combining importance-weighted keyframe selection, domain-adapted graph reconstruction, and iterative process summarization—produces the most faithful process traces and consistently delivers the highest-quality generated materials.

To illustrate the role of each component more concretely, we analyze a representative example: generating a \emph{raw–beef} material. This example is diagnostically useful because high-quality results require (1) multi-scale feature construction, (2) correct node-graph topology, and (3) coherent cross-channel reasoning across color, roughness, bump, and displacement. Even small parsing failures therefore produce large and visually noticeable artifacts. Figures~\ref{fig:VUMablation_audio}–\ref{fig:VUMablation_full} show qualitative results for each ablation.

\textbf{Audio-Only.}  
With only narration available, the system has no access to graph structure or spatial cues present in the tutorial frames. As shown in Figure~\ref{fig:VUMablation_audio}, the reconstructed graph collapses into a minimal structure driven almost entirely by a single reused ColorRamp mask, which is incorrectly repurposed across color, roughness, bump, and displacement. The rendered output becomes a flat two-tone pattern with no fat content, strand flow, or multi-scale variation. Audio alone does not describe the topology, grouping, or numeric structure required for procedural reconstruction.

\textbf{Random Keyframe Sampling} (Figure~\ref{fig:VUMablation_randomsample}).  
Using visual frames but selecting them uniformly at random exposes many relevant nodes but rarely captures the structural relationships between them. Important edges and parameter interactions simply never appear in the sampled frames. As shown in Figure~\ref{fig:VUMablation_randomsample}, the system assembles the correct primitives (Noise, Musgrave, Voronoi) but connects them incorrectly: displacement becomes disconnected from color reasoning, and the intended directional warp collapses into an isotropic bump field. The result resembles a lumpy sausage interior rather than raw beef. This demonstrates that which frames are selected is as important as having frames at all.

\textbf{GPT-5 Frame Parsing}.  
Replacing the domain-adapted detectors with generic GPT-5 vision parsing yields noisier node states. GPT-5 frequently misreads sliders, color ramps, and numeric parameters—particularly in complex nodes such as ColorRamp or Musgrave. As a result (Figure~\ref{fig:VUMablation_gpt}), frequency values become exaggerated, color-ramp thresholds collapse to black, and vector outputs are misinterpreted as scalar fields. The resulting material resembles an over-sharpened sponge rather than meat. This highlights the need for domain-adapted parsing capable of reliably extracting Blender-specific parameters.

\textbf{No Self-Refinement in Process Summarization}.  
When keyframes and graph reconstruction are correct but the summarizer compresses the verbose process trace in a single pass, subtle structural dependencies are lost. As shown in Figure~\ref{fig:VUMablation_norefine}, the graph contains the correct families of nodes (fat masks, noise layers, fiber masks) but their relationships become mis-sequenced or collapsed. The rendering resembles ground meat—plausible in isolation but lacking the anisotropic strand-aligned structure of raw beef. Iterative refinement is required to preserve long-range dependencies across channels.

\textbf{Full Pipeline}.  
Only the complete pipeline—importance-weighted keyframes, domain-adapted graph reconstruction, accurate color and parameter extraction, and iterative process summarization—successfully reconstructs the procedural logic of the tutorial. As shown in Figure~\ref{fig:VUMablation_full}, the recovered graph captures the intended multi-scale structure: a warped vector field for strand direction, a mid-frequency Musgrave mask for elongated fibers, Voronoi fat specks modulating color and roughness, micro-noise for fine grain, and consistent cross-channel connections tying these elements together. The resulting render exhibits anatomically plausible marbling, strand flow, and shading, demonstrating that all stages of the pipeline are required to recover the full procedural structure of expert workflows.

\subsection{Refinement}
\begin{figure*}[t]
    \centering
    \begin{overpic}[width=0.9\textwidth]{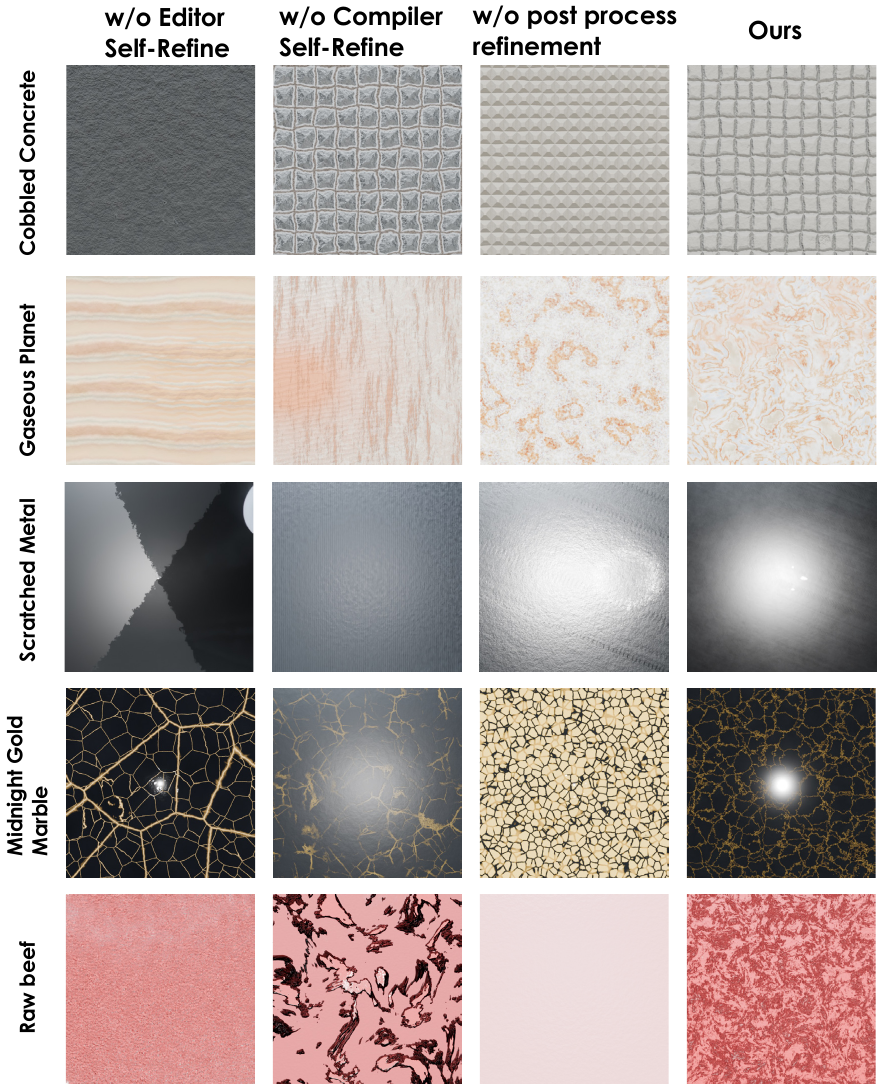}
    \end{overpic}
    \caption{\textbf{Ablation of refinement strategies :} We evaluate the impact of three refinement components on text-to-material generation: removing self-refinement from the Editor, removing self-refinement from the Compiler, and omitting post-process refinement (BlenderAlchemy-style iterative improvement). Across five representative materials, removing either self-refinement stage leads to noticeable structural and perceptual degradation, while skipping post-process refinement reduces realism and prompt alignment. Our full pipeline (rightmost column), which includes all refinement stages, consistently produces the most faithful and visually coherent materials.
    }
    \label{fig:refinement_ablation}
\end{figure*}

Refinement plays a key role in improving generation quality in \ma. We study two forms of refinement: (1) Self-Refine\cite{madaan2023self}, where an LLM critiques and revises its own output, and (2) iterative material-graph refinement using BlenderAlchemy\cite{huang2024blenderalchemy}. Motivated by the gains observed from self-refine in the \summarizer, we apply the same strategy to both the \editor\ and \com\ and evaluate their individual contributions. Figure~\ref{fig:refinement_ablation} shows qualitative results across five text-to-material examples. Removing self-refine from either the \editor\ or \com\ causes a clear drop in fidelity and coherence. Although the raw outputs from the \com\ are already strong, subsequent iterative refinement further improves realism and prompt alignment, demonstrating complementary benefits between LLM-based self-correction and post-hoc graph optimization.

\subsection{Decompiler}
\begin{figure}[t]
    \centering
    \begin{overpic}[width=\linewidth]{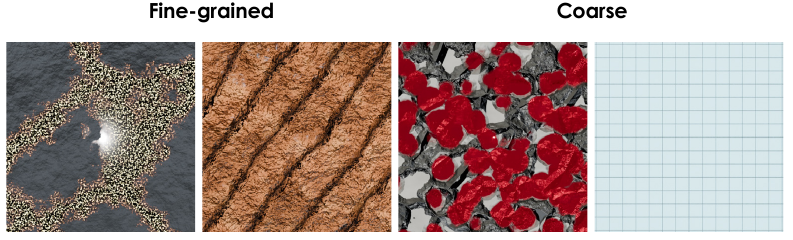}
    \end{overpic}
    \caption{\textbf{Ablation of the \dcom :} Editing results on the same prompts as in the main paper. Without the \dcom\ (results shown here), the editor receives only the raw input graph and produces weak or semantically drifting edits. With the \dcom\ (results shown in main paper Figure 7), converting the graph into a process trace enables stable, aligned, and higher-quality edits.
    }
    \label{fig:decompiler_ablation}
\end{figure}
The \dcom\ converts an input material graph $g_i$ into a process trace $p_i$, allowing the \editor\ to operate in the same modality as the retrieved process traces. Using a small bank of exemplar (graph, process) pairs, it aligns substructures in $g_i$ with analogous patterns in these exemplars to reconstruct a plausible authoring process. Without this step, the \editor\ must edit graphs directly, which leads to unstable edits and semantic drift. Figure~\ref{fig:decompiler_ablation} shows this effect on the \emph{same editing setup and prompts used in the main paper}: bypassing the \dcom\ and supplying the initial graph directly to the \editor\ produces noticeably weaker and less controlled edits compared to our full pipeline.

\section{Comparison to PhotoMat}

\begin{figure}[t]
    \centering
    \begin{overpic}[width=\textwidth]{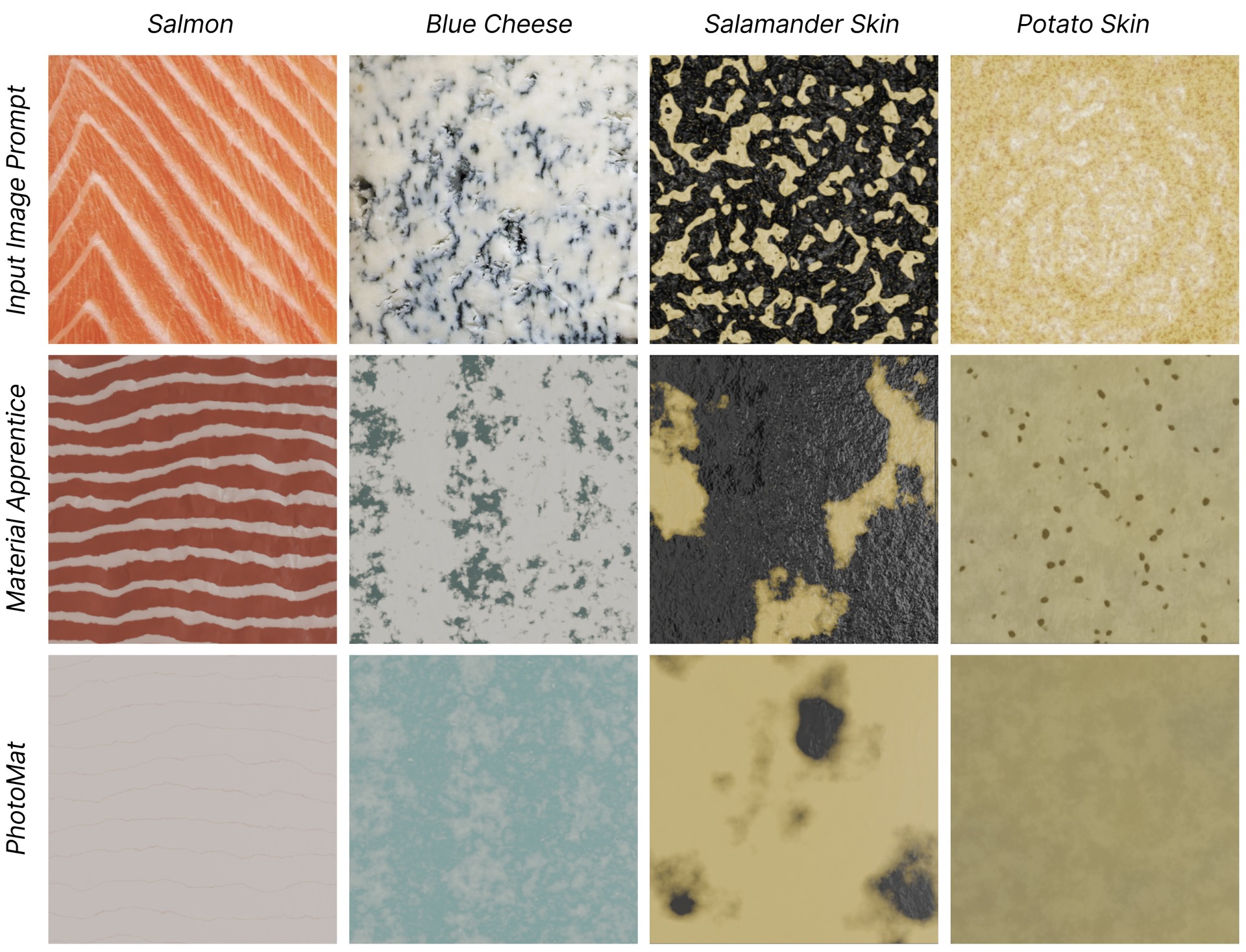}
    \end{overpic}
    \vspace{-0.7cm}
    \caption{\small Qualitative comparison with PhotoMat~\cite{zhou2023photomat}. Given an input reference image (top row), \ma\ generates procedural materials that better reproduce the characteristic structural patterns of the target materials. In contrast, PhotoMat—despite capturing approximate color statistics—often produces overly smooth or structurally simplified textures due to its purely image-based generation paradigm.}
    \vspace{-0.3cm}
    \label{fig:photomat_comp}
    \vspace{-0.1cm}
\end{figure}

We provide qualitative comparisons with PhotoMat~\cite{zhou2023photomat}, a state-of-the-art \emph{non-procedural} texture generation method based on diffusion models. Unlike \ma, which generates procedural shader graphs, PhotoMat directly synthesizes texture maps from image inputs. Figure~\ref{fig:photomat_comp} shows representative results on several materials. While PhotoMat often captures coarse color statistics, it frequently struggles to reproduce the structured, hierarchical patterns present in many materials. In contrast, \ma\ produces outputs that better preserve directional structures and characteristic construction patterns (e.g., the layered fibers in \textit{Salmon} or the clustered patterns in \textit{Salamander Skin}), highlighting the advantages of process-based procedural generation.

\section{Additional Results}
\begin{figure*}[t]
    \centering
    \begin{overpic}[width=\textwidth]{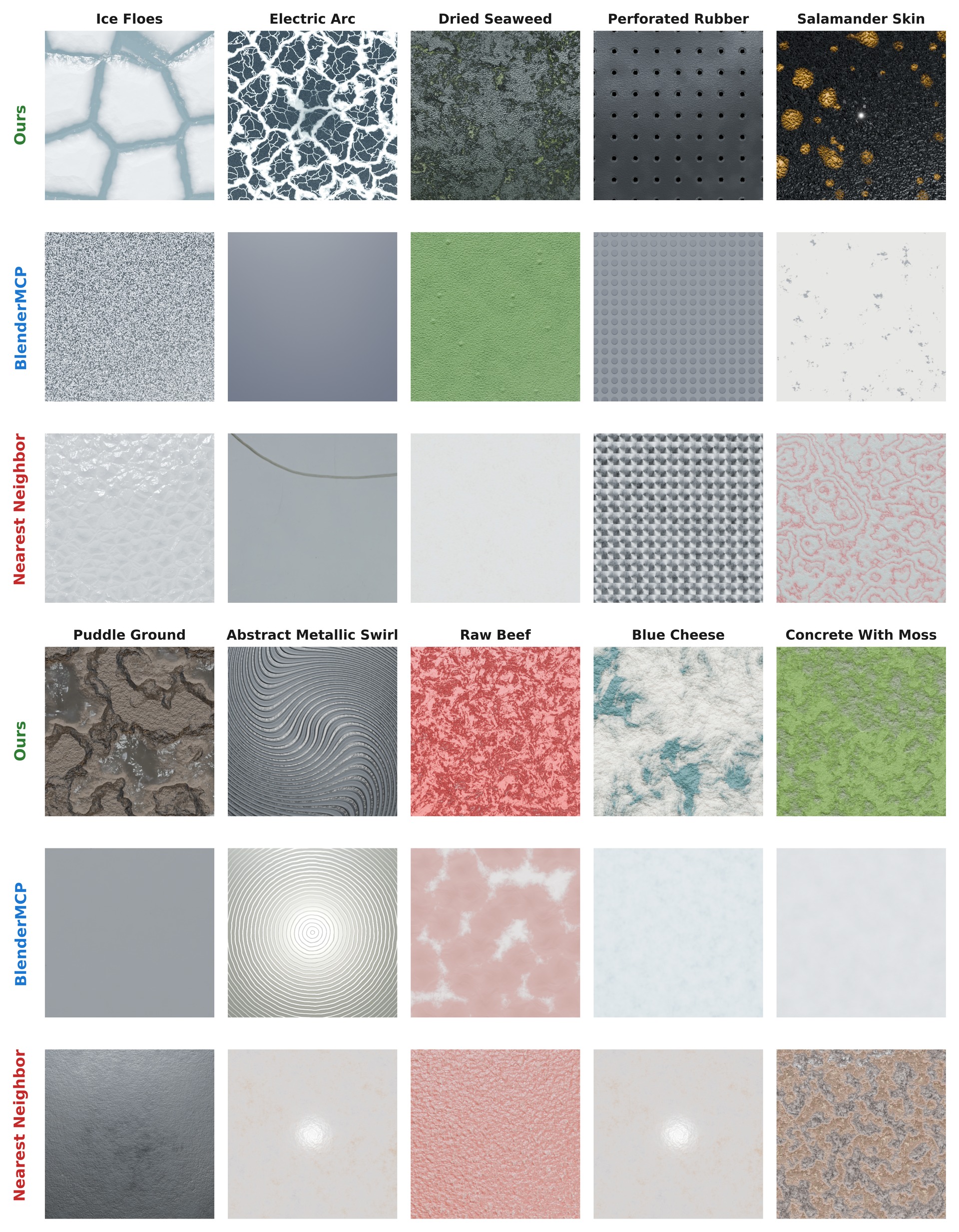}
    \end{overpic}
    \caption{Example showing the Text to material Generation for 10 prompts.
    }
    \label{fig:text2mat_supp1}
\end{figure*}

\begin{figure*}[t]
    \centering
    \begin{overpic}[width=\textwidth]{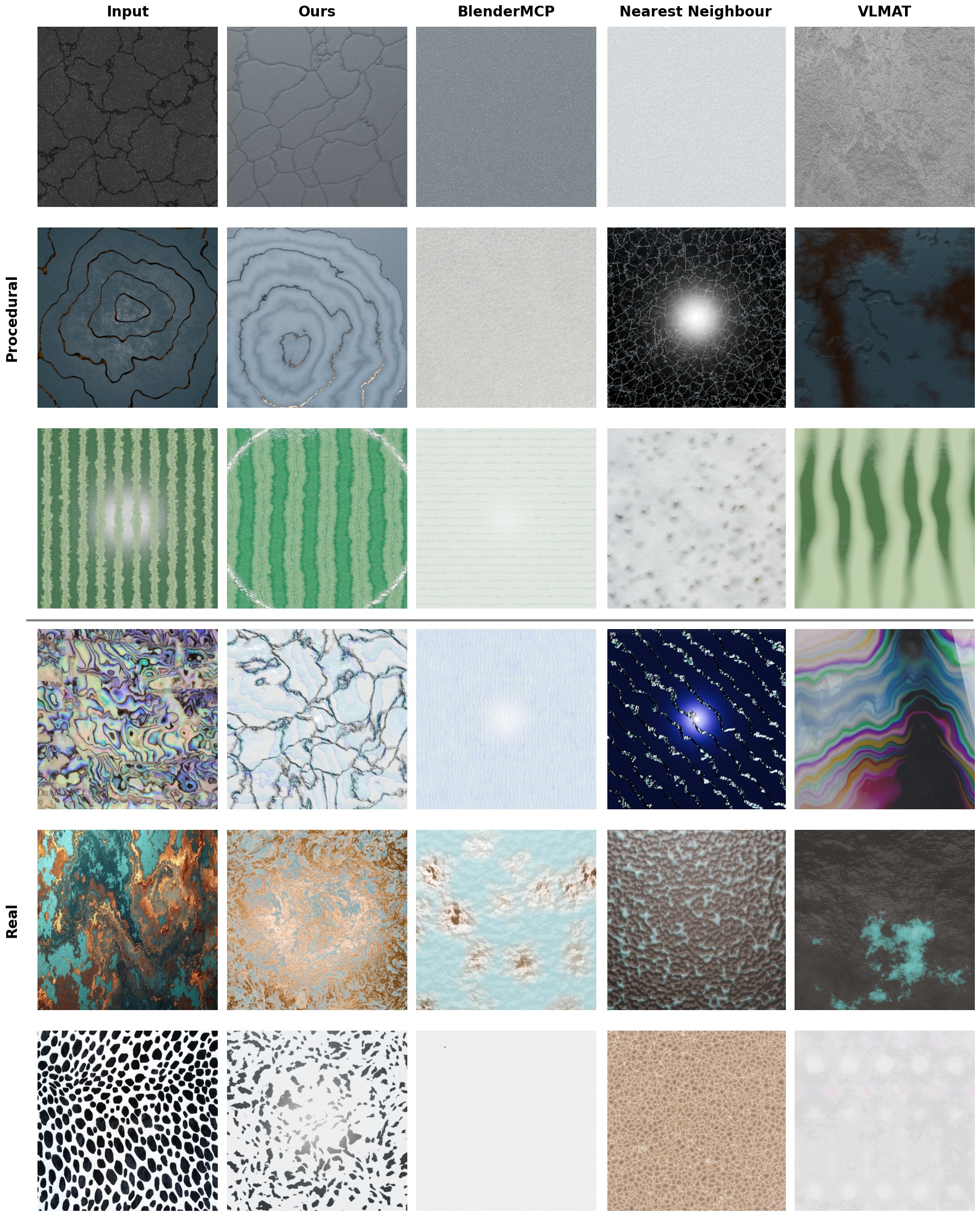}
    \end{overpic}
    \caption{\textbf{Image conditioned Material Generation.} Images include whethered asphalt, midnight gold marble, watermelon, abalone shell sheet, copper platina, dalmatian skin. Our method better captures micro and macro details specified via the image reference.
    }
    \label{fig:img_to_mat_sup1}
\end{figure*}

\begin{figure*}[t]
    \centering
    \begin{overpic}[width=\textwidth]{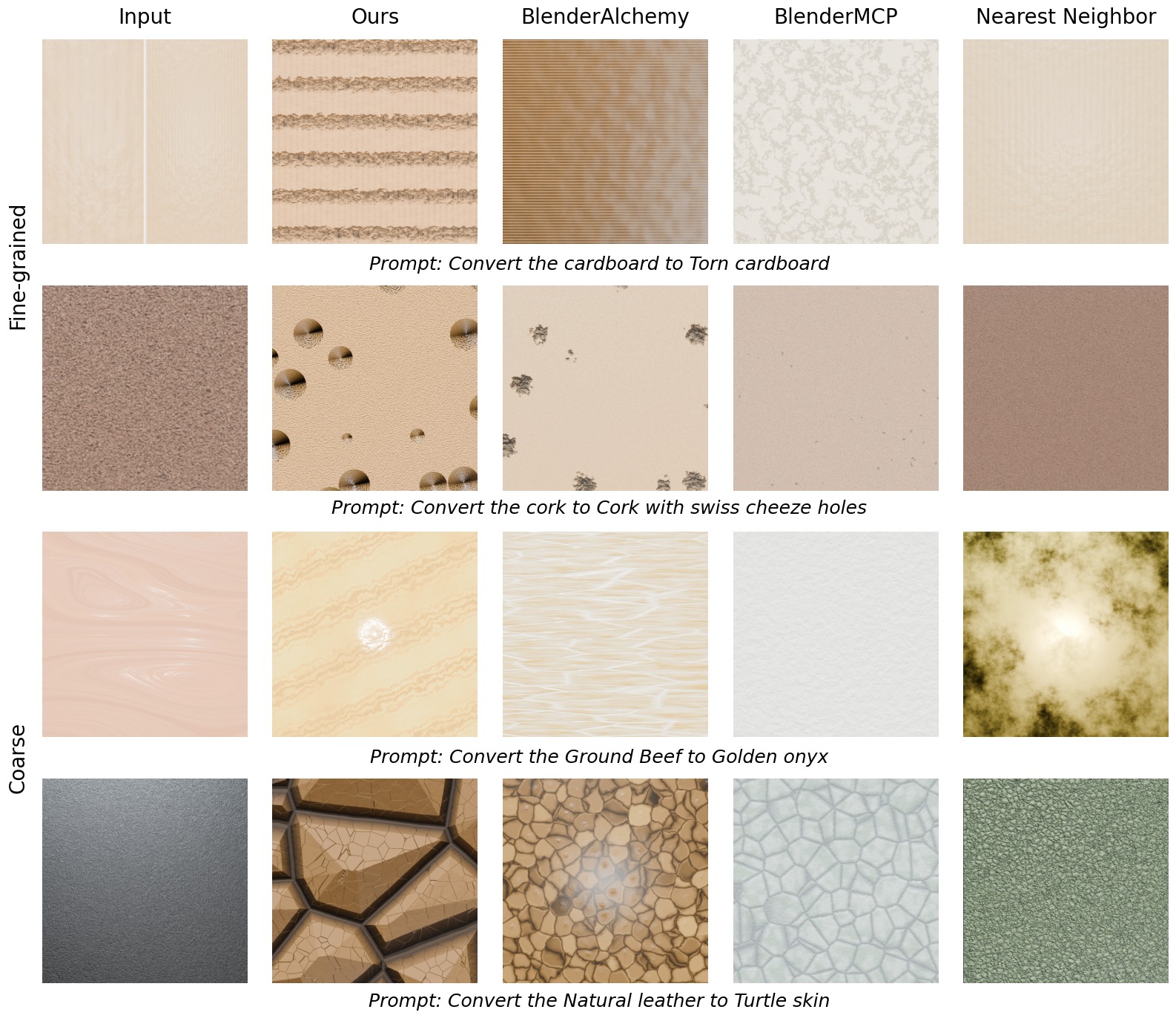}
    \end{overpic}
    \caption{Fine and Coarse editing on a base material given certain prompts.
    }
    \label{fig:editing_sup}
\end{figure*}

Additional qualitative results for text conditioned generation, image conditioned generation and text based editing are shown in Figures \ref{fig:text2mat_supp1}, \ref{fig:img_to_mat_sup1}, \ref{fig:editing_sup} respectively. 

\section{Limitations}
\ma\ is the first system to explicitly leverage tacit expertise in procedural material creation to synthesize new materials, but several limitations remain.
\begin{figure}[t]
    \centering
    \begin{overpic}[width=0.8\linewidth]{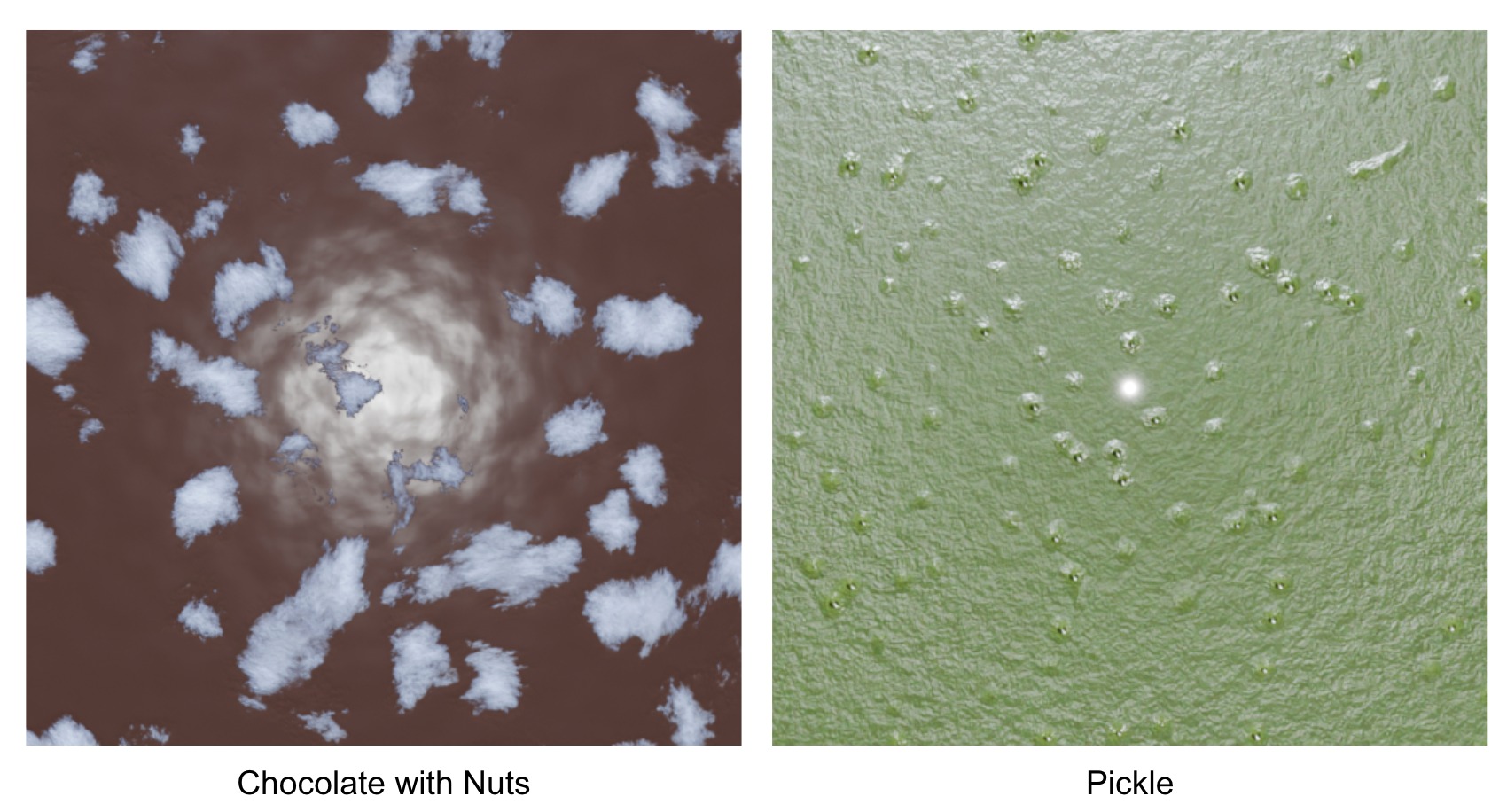}
    \end{overpic}
    \caption{\textbf{Limitations :} \ma\ stuggles to generate materials requiring more than 25 nodes. Example shown here for text to material generation for chocolate with nuts and pickle, both of which require significant number of shader nodes to express adequate realism.
    }
    \label{fig:lim}
\end{figure}

\begin{figure}[h]
    \centering
    \begin{overpic}[width=\textwidth]{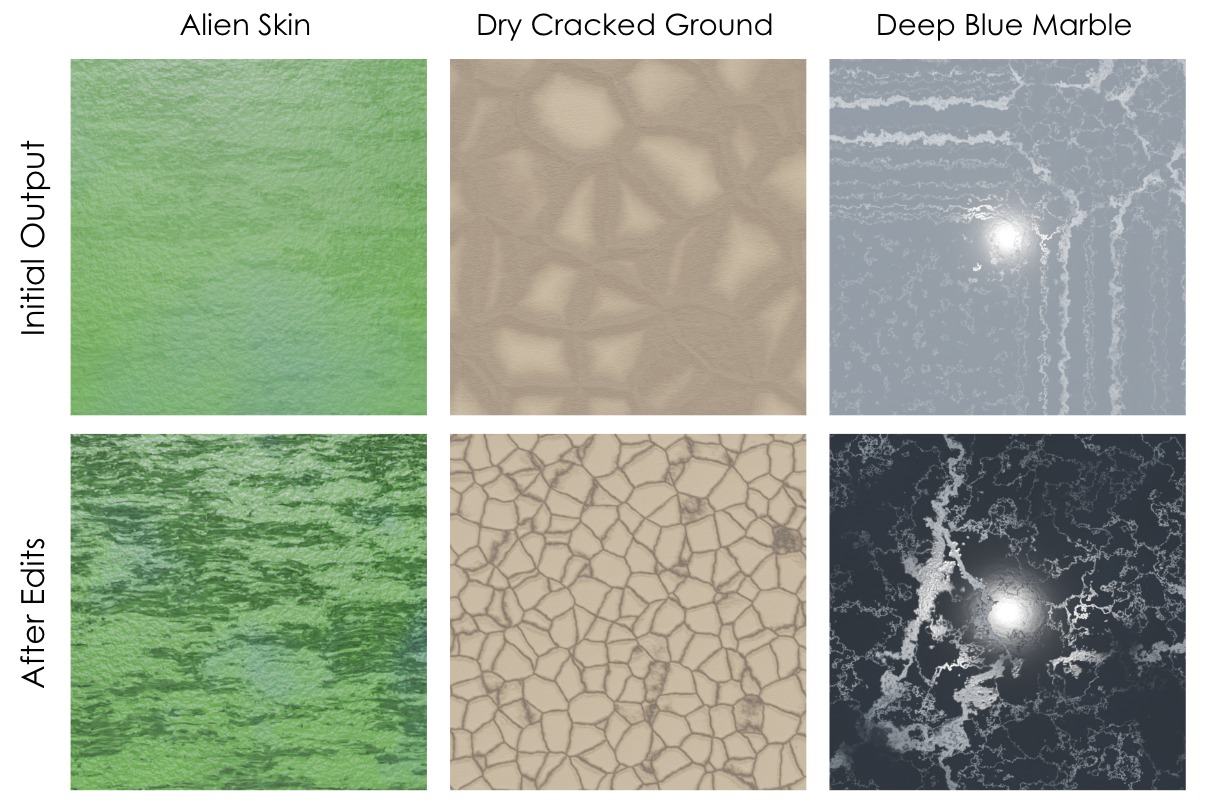}
    \end{overpic}
    \caption{\small\textbf{Limitations and simple refinements.}
    Our method can occasionally produce materials that lack clear structural detail or exhibit imperfect texture organization. 
    The top row shows the initial outputs generated by the pipeline for three example materials (Alien Skin, Dry Cracked Ground, and Deep Blue Marble). 
    While the overall color and coarse appearance are captured, important meso-scale structures such as cracks, veins, or directional patterns may be weak or poorly formed. 
    The bottom row shows the same materials after applying small parameter edits to the generated procedural graph. 
    These lightweight adjustments recover sharper structural features and more realistic appearance, indicating that the generated materials are often close to a correct solution but may require minor refinement.}
    \label{fig:Limitations_fig}
\end{figure}

First, expert-made materials often fall into two extremes: highly elaborate graphs (see Fig\ref{fig:lim}) with many interacting layers, or extremely concise graphs that use a few nodes to express rich structure. \ma\ struggles in both regimes. Large graphs introduce too many degrees of freedom for reliable synthesis, while compact, elegant graphs reflect artistic intuition developed over years—intuition that cannot be captured from a small number of demonstrations \ref{fig:lim}.
Second, as shown in Figure~\ref{fig:Limitations_fig}, while \ma\ created materials are often structurally close, they sometimes lack adequate parameter tuning which may require some human inputs to achieve desired quality.
Lastly, \ma\ depends on a powerful but slow and expensive backbone LLM (GPT-5) for both video understanding and process synthesis in a single forward pass. Future work could explore more efficient multi-agent or modular architectures that decompose the task—e.g., separating video parsing, structural reasoning, and graph generation—allowing these components to be handled by smaller, faster, and more accessible models.


\begin{figure}[t]
    \centering
    \begin{overpic}[width=\textwidth]{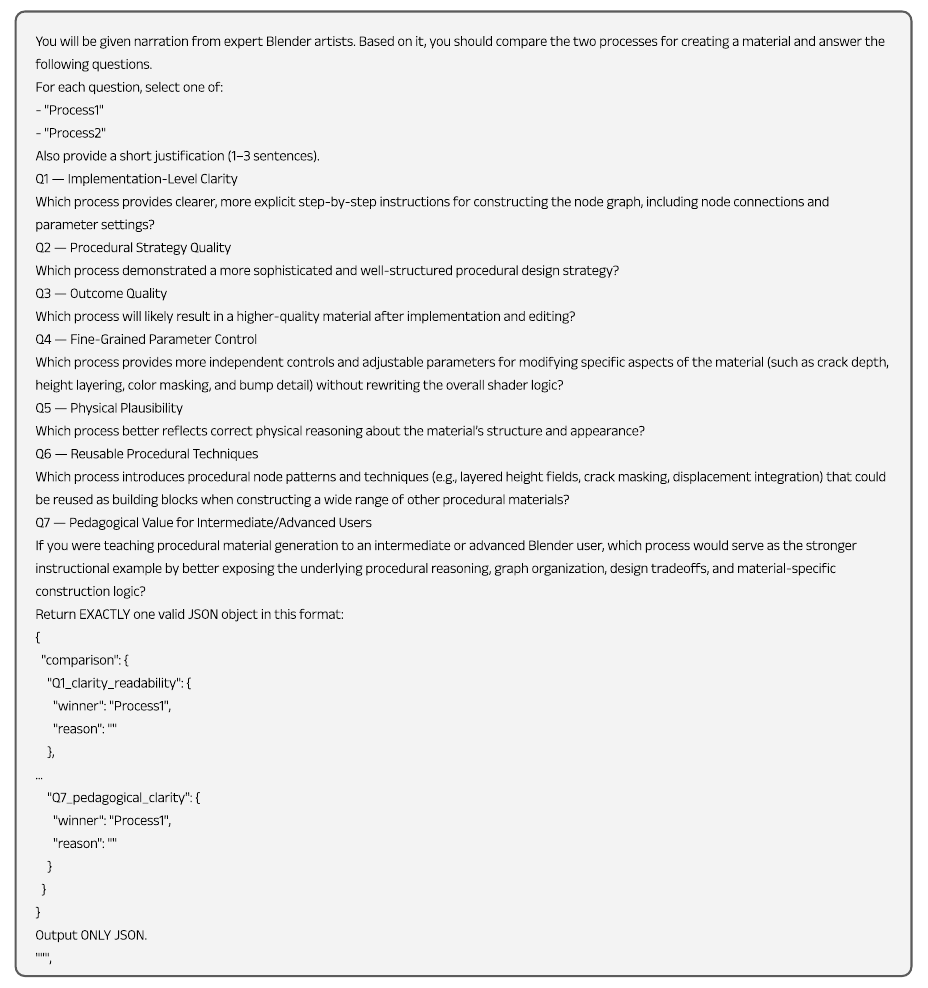}
    \end{overpic}
    \caption{\small Prompt used to analyze artist narrations and perform the structured forced-choice evaluation across the defined criteria (Q1–Q7).}
    \label{fig:question_prompt}
\end{figure}

\begin{figure}[t]
    \centering
    \begin{overpic}[width=\textwidth]{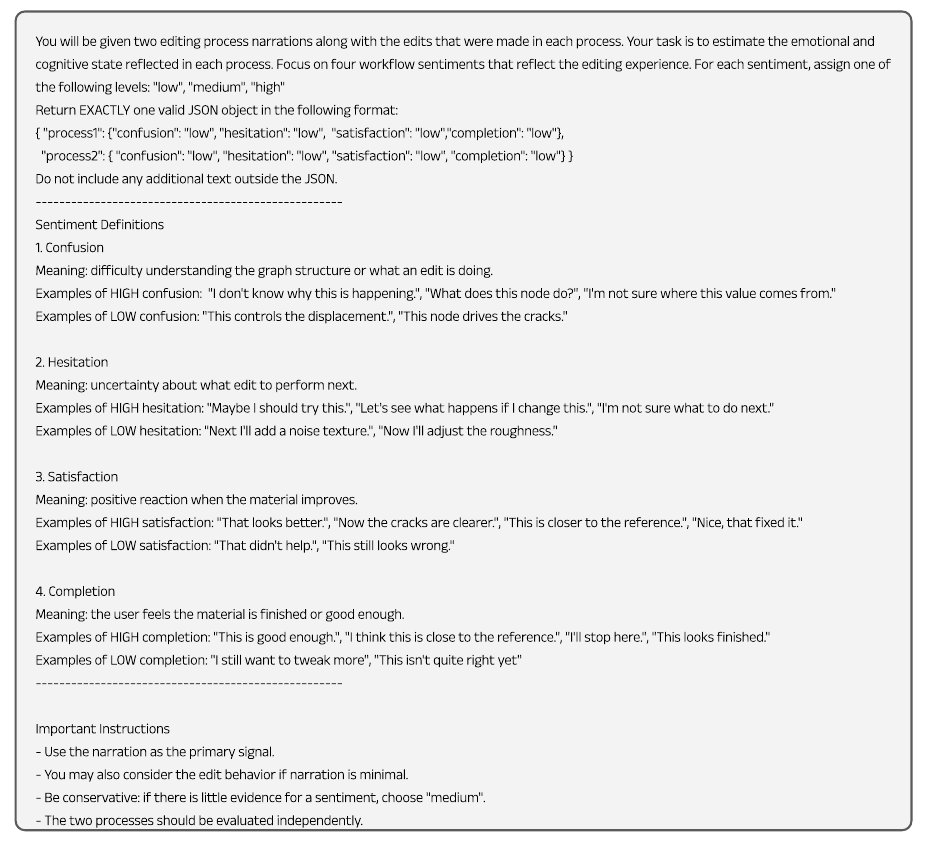}
    \end{overpic}
    \caption{\small Prompt used for sentiment analysis of the editing narrations, identifying signals such as confusion, hesitation, satisfaction, and completion.}
    \label{fig:sentiment_prompt}
\end{figure}

\begin{figure}[t]
    \centering
    \begin{overpic}[width=\textwidth]{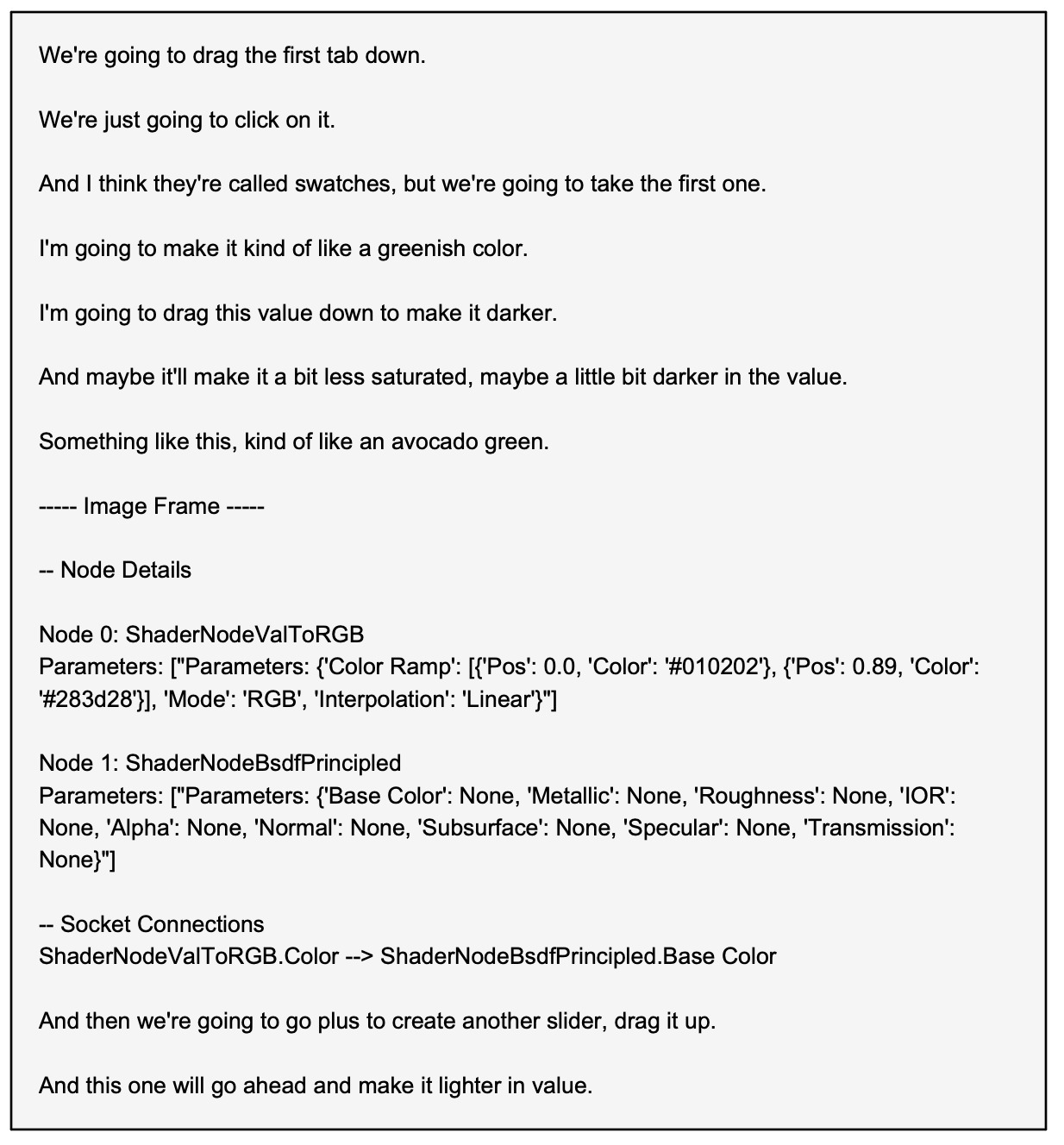}
    \end{overpic}
    \caption{\small Example raw narration transcript extracted from a tutorial video (Dinosaur skin material). The transcript captures the full verbal explanation provided by the artist and forms the initial verbose procedural trace before summarization.}
    \label{fig:transcript}
\end{figure}

\begin{figure}[t]
    \centering
    \begin{overpic}[width=\textwidth]{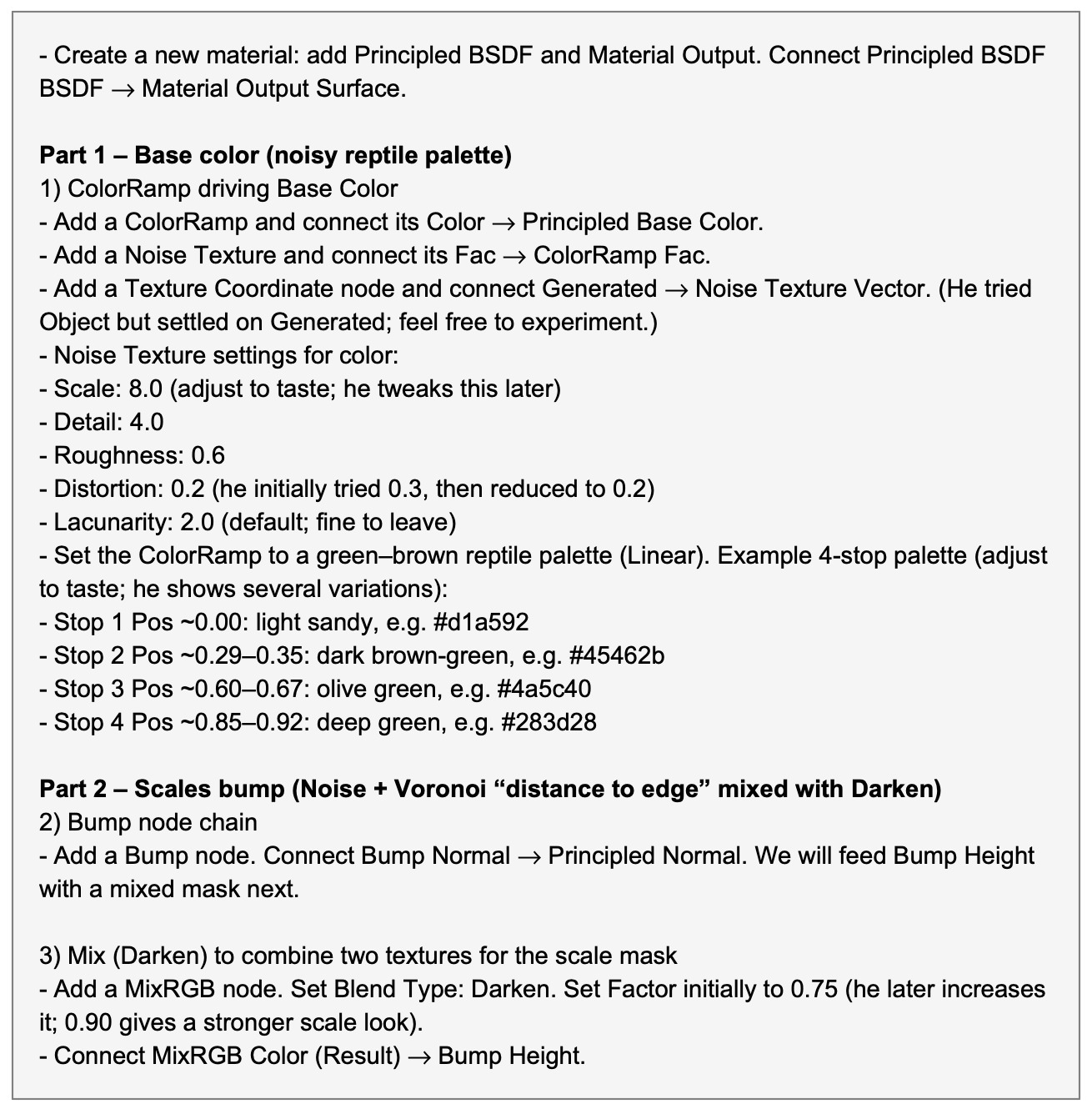}
    \end{overpic}
    \vspace{-0.7cm}
    \caption{\small Example summarized process trace generated by the \summarizer\ for the Dinosaur skin material. The summarization condenses the verbose transcript and reconstructed graph states into a textual process trace suitable for conditioning the \ps.}
    \vspace{-0.3cm}
    \label{fig:procedure}
    \vspace{-0.1cm}
\end{figure}
\end{document}